\pdfoutput=1

\documentclass[11pt]{article}
\usepackage[a4paper,vmargin={30mm,30mm},hmargin={35mm,35mm}]{geometry}
\usepackage{amsmath}
\usepackage{amsfonts}
\usepackage{graphicx}
\usepackage{amsthm}
\usepackage{parskip}
\usepackage{setspace}
\usepackage{times}
\usepackage{makecell}
\usepackage{array}

\usepackage[table]{xcolor}
\usepackage{tabularx}
\usepackage{multirow}
\usepackage{booktabs}
\usepackage{enumitem}
\usepackage[numbers]{natbib}
\usepackage{hyperref}
\usepackage{url}
\usepackage{graphicx}
\usepackage{subcaption}
\usepackage[T1]{fontenc}
\usepackage[utf8]{inputenc}
\usepackage{float}
\usepackage{microtype}

\usepackage{xcolor}
\usepackage{cleveref}

\definecolor{CBblue}{RGB}{0, 107, 164}
\definecolor{CBorange}{RGB}{255, 128, 14}
\definecolor{CBgray}{RGB}{89, 89, 89}
\definecolor{CByellow}{RGB}{171, 171, 42}
\definecolor{CBpurple}{RGB}{137, 61, 86}
\definecolor{CBgreen}{RGB}{27, 158, 119}
\definecolor{CBpink}{RGB}{215, 48, 39}

\usepackage{tcolorbox}
\usepackage{fontawesome}

\tcbuselibrary{skins}
\newtcolorbox{challsumm}{
  enhanced,
  colframe=red!70!black,
  colback=red!10!white,
  coltitle=white,
  fonttitle=\bfseries,
  sharp corners,
  boxrule=1pt,
  drop fuzzy shadow
}

\newtcolorbox{riskfactor}[1]{
  enhanced,
  colframe=blue!70!black,
  colback=blue!10!white,
  coltitle=white,
  fonttitle=\bfseries,
  title={\faWarning\hspace{0.5em}\textbf{#1}},
  sharp corners,
  boxrule=1pt,
  drop fuzzy shadow
}

\newtcolorbox{burden}[1]{
  enhanced,
  colframe=orange!70!black,
  colback=orange!10!white,
  coltitle=white,
  fonttitle=\bfseries,
  title={\faStethoscope\hspace{0.5em}\textbf{#1}},
  sharp corners,
  boxrule=1pt,
  drop fuzzy shadow
}

\newtcolorbox{intervention}[1]{
  enhanced,
  colframe=green!70!black,
  colback=green!10!white,
  coltitle=white,
  fonttitle=\bfseries,
  title={\faPlusCircle\hspace{0.5em}\textbf{#1}},
  sharp corners,
  boxrule=1pt,
  drop fuzzy shadow
}

\newtcolorbox{constraintsumm}{
  enhanced,
  colframe=orange!70!black,
  colback=orange!10!white,
  coltitle=white,
  fonttitle=\bfseries,
  sharp corners,
  boxrule=1pt,
  drop fuzzy shadow
}

\Crefname{section}{Sec.}{Secs.}
\Crefname{equation}{Eq.}{Eqs.}
\Crefname{figure}{Fig.}{Figs.}
\Crefname{tabular}{Tab.}{Tabs.}

\usepackage{longtable}

\usepackage{pifont}

\usepackage{tikz}
\usepackage{fontawesome}

\definecolor{darkblue}{rgb}{0, 0, 0.5}
\hypersetup{colorlinks=true, citecolor=darkblue, linkcolor=darkblue, urlcolor=darkblue}

\RequirePackage{caption}
\DeclareCaptionFont{10pt}{\fontsize{10pt}{12pt}\selectfont}
\usepackage{etoolbox}
\patchcmd{\thebibliography}{\section*{\refname}}{}{}{}

\begin{document}
{\Large \bf Evaluating Large Language Models for Public Health \\Classification and Extraction Tasks}

\linespread{1}

{\small 
Joshua Harris$^{1, \dagger}$,
Timothy Laurence$^{1}$,
Leo Loman$^{1}$,
Fan Grayson$^{1}$,
Toby Nonnenmacher$^{1}$,
Harry Long$^{1}$,
Loes WalsGriffith$^{1}$,
Amy Douglas$^{1}$,
Holly Fountain$^{1}$,
Stelios Georgiou$^{1}$,
Jo Hardstaff$^{1}$,
Kathryn Hopkins$^{1}$,
Y-Ling Chi$^{1}$,
Galena Kuyumdzhieva$^{1}$,
Lesley Larkin$^{1}$,
Samuel Collins$^{1}$,
Hamish Mohammed$^{1}$,
Thomas Finnie$^{1}$,
Luke Hounsome$^{1}$,
Michael Borowitz$^{1}$,
and Steven Riley$^{1}$
}
{\small \begin{enumerate}
\item UK Health Security Agency 
\end{enumerate}}
\renewcommand*{\thefootnote}{\fnsymbol{footnote}}
\footnotetext[2]{Corresponding author: joshua.harris@ukhsa.gov.uk}
\renewcommand*{\thefootnote}{\arabic{footnote}}

\begin{abstract}
Advances in Large Language Models (LLMs) have led to significant interest in their potential to support human experts across a range of domains, including public health. In this work we present automated evaluations of LLMs for public health tasks involving the classification and extraction of free text. We combine six externally annotated datasets with seven new internally annotated datasets to evaluate LLMs for processing text related to: health burden, epidemiological risk factors, and public health interventions. We evaluate eleven open-weight LLMs (7-123 billion parameters) across all tasks using zero-shot in-context learning. We find that Llama-3.3-70B-Instruct is the highest performing model, achieving the best results on 8/16 tasks (using micro-F1 scores). We see significant variation across tasks with all open-weight LLMs scoring below 60\% micro-F1 on some challenging tasks, such as \textit{Contact Classification}, while all LLMs achieve greater than 80\% micro-F1 on others, such as \textit{GI Illness Classification}. For a subset of 11 tasks, we also evaluate three GPT-4 and GPT-4o series models and find comparable results to Llama-3.3-70B-Instruct. Overall, based on these initial results we find promising signs that LLMs may be useful tools for public health experts to extract information from a wide variety of free text sources, and support public health surveillance, research, and interventions.       
\end{abstract}

\begin{figure}[H]
    \hspace*{0.2cm}
    \centering
    \begin{subfigure}[b]{0.42\textwidth}
        \centering
        \includegraphics[width=\textwidth]{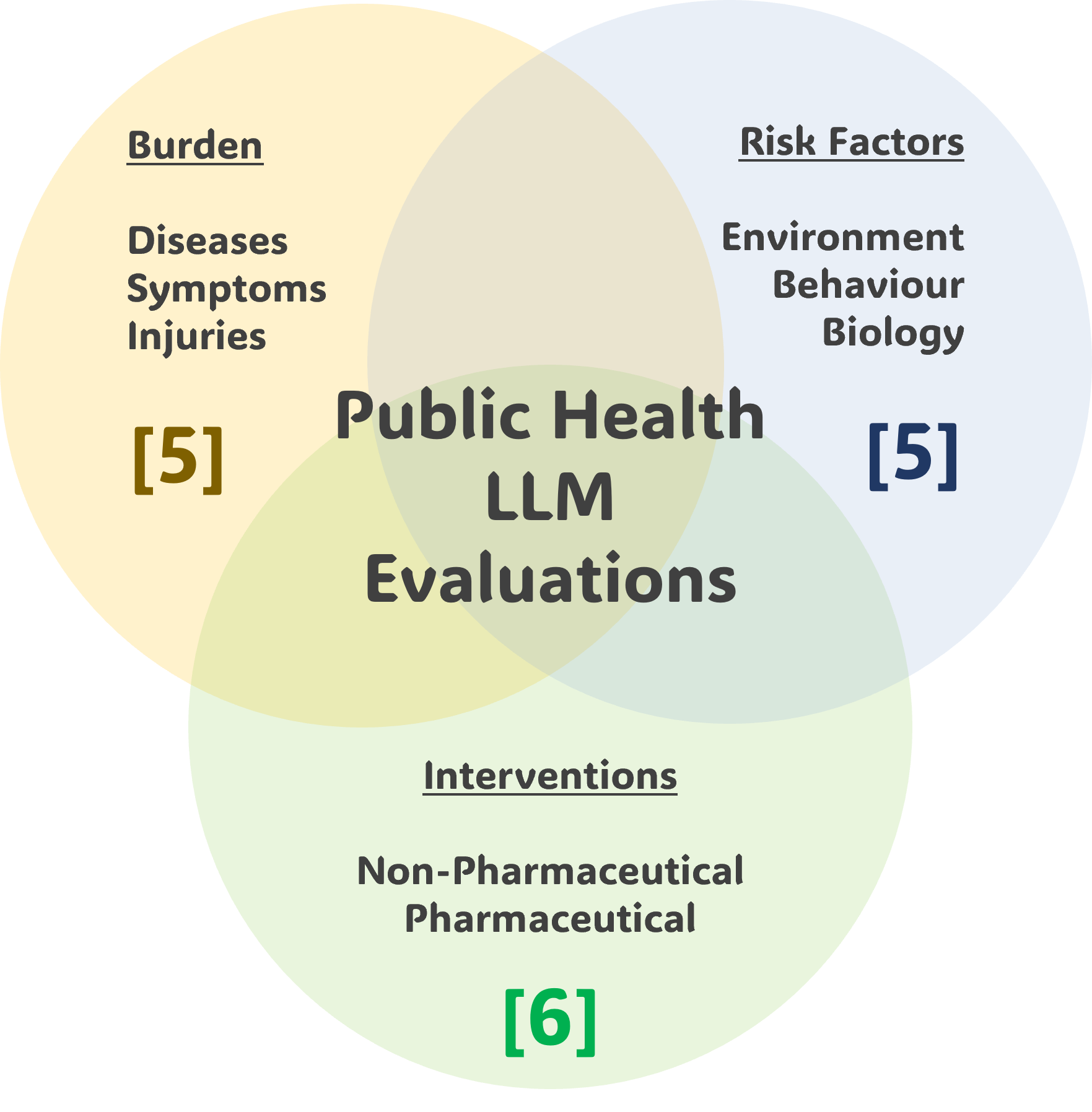}
    \end{subfigure}
    \hfill 
    \begin{subfigure}[b]{0.55\textwidth}
    \hspace*{0.5cm}
        \centering
        \includegraphics[width=\textwidth]{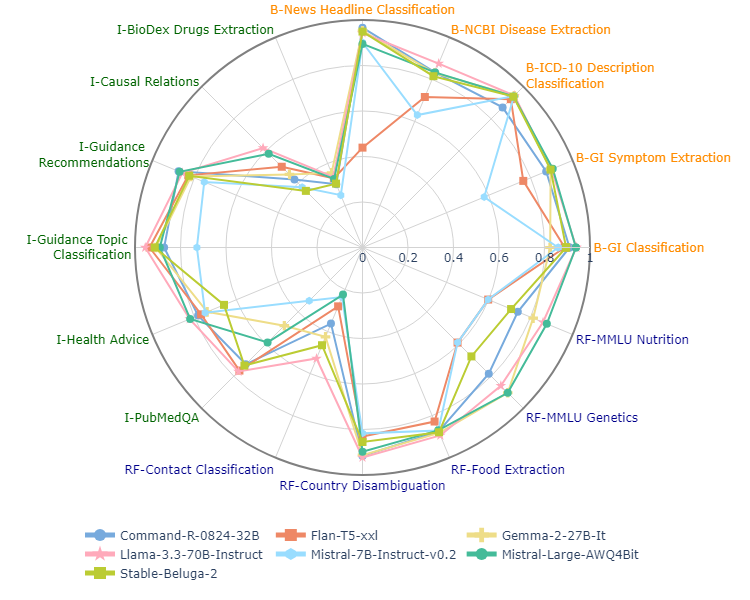}
    \end{subfigure}
 \captionsetup{font=small}
    \caption{\textbf{Public Health Large Language Model (LLM) Evaluation Areas [Number of Evaluations] and Task Evaluation Micro-F1 Scores by Model}. (Left) We divide public health free text processing into three sub-domains: (1) burden, such as reports of disease symptoms, cases, morbidity, or mortality; (2) risk factors, such as environmental, behavioural, or biological contributors; (3) interventions, pharmaceutical and non-pharmaceutical. (Right) Evaluation results (micro-F1 scores) for seven open-weight LLM architectures (highest performing model evaluated from each) across the 16 tasks using zero-shot prompting.}
    \label{fig:main_fig}
\end{figure}

\clearpage

\section{Introduction}
There have been rapid improvements in the ability of Large Language Models (LLMs)\footnote{For background,~\citet{zhao2023survey} and ~\citet{kaddour2023challenges} are recent surveys on LLMs and their applications.} to perform a broad range of text processing tasks~\citep{bubeck2023sparks, openai2023gpt4, medpalm2}. This has led to significant interest in applying them to support experts across a range of domains~\cite{eloundou2023gpts}, including public health~\cite{ukhsa_ai_board, cdcArtificialIntelligence, baclic2020artificial}.

Previous work has often demonstrated considerable variation in how different LLMs perform on a given task, and that a given LLM's performance is highly dependent on the type and nature of the task involved~\citep{chen2024rarebench, chen2024benchmarking, xie2024finben, fei2023lawbench, helm_website}. Therefore, to understand the potential effective application of LLMs within public health an important prerequisite is developing domain specific evaluations that are representative of the tasks, free text, and knowledge that human experts regularly come across. Here, our overarching aim is to understand what factors determine variations in performance, whether that be: the LLM, the nature of the task, the type of free text, or the specific implementation details.   

We evaluate LLMs across a wide range of public health tasks and free text. In this initial work, using the definitions provided by \citet{chang2023survey}, we focus on \textit{Automated Evaluation} (excluding LLM-as-a-Judge~\citep{zheng2023judging}) of \textit{Natural Language Understanding} (NLU) tasks (e.g classification and inference). We also focus solely on evaluating LLM in-context learning (prompting) approaches. 

These initial evaluations enable us to start assessing LLMs for potential use in public health in an automated way, including: (1) continually assessing new and existing private and open-weight LLMs for their potential applicability to public health, (2) identifying specific areas of public health where LLMs could potentially be applied, (3) providing a baseline for fine-tuning public health specific LLMs in the future. 

We see this work as an important first step to understanding the potential of LLMs to perform public health free text processing tasks. Further research is needed to investigate long form \textit{Natural Language Generation} (NLG)~\cite{chang2023survey} tasks, with evaluation by public health experts, as well as in-depth studies on specific use-cases and potential issues such bias.  

\section{Methods}

Evaluation of LLMs is a broad and rapidly growing field of research~\cite{chang2023survey} ranging from very general assessments of capabilities or intelligence~\cite{bubeck2023sparks} to very task specific performance results~\cite{samaan2023assessing} (\Cref{fig:venn_eval_fig}). We focus on domain and task specific evaluations of LLMs within public health and provide a review of the relevant literature in \Cref{lit_review_full}. 

The appropriate evaluation methodology largely depends on three factors: level of generality (see \Cref{lit_generality}), types of task (see \Cref{lit_task_type}), and outcome of interest (see \Cref{lit_outcome}). For our initial evaluations of LLMs within public health, we aim to provide a high level assessment of LLM performance over a broad range of tasks, models, and sub-domains. To enable this we focus on classification and extraction NLU tasks that can be assessed using automated evaluation approaches. By collecting representative internal data and annotations in collaboration with public health experts, we aim to also provide early evidence of potential task specific performance in individual areas. Specifically we use:

\begin{enumerate} [label=\bfseries\arabic*.,leftmargin=*,align=left]
     \item \textit{7 New Annotated Datasets} - We collect and manually annotate seven datasets with public health specific annotations using a combination of internal, synthetic, and external free text sources. 
    \item \textit{Existing Datasets and Literature} - We identify and include six evaluation datasets from existing work that are applicable to public health. To inform and develop our public health evaluations we also review the literature on relevant evaluations in related domains, such as Medicine. We provide a detailed overview of these external results and datasets in \Cref{lit_evals} and \Cref{sec_tasks}. 
    \item \textit{16 Public Health Specific Evaluation Tasks} - In total we bring together 16 classification and extraction evaluation tasks across three sub-domains of public health, see \Cref{fig:main_fig}.
    \item \textit{11 Open-weight LLMs Evaluated and GPT-4 Series Comparisons} - We deploy and assess eleven open-weight models ranging from 7-123 billion parameters across all tasks. For a subset of 11 tasks, we also evaluate three GPT-4 series models (turbo, 4o, and 4o-mini) in order to understand how open-weight model performance compares to some of the highest performing private models on our public health tasks.
\end{enumerate}

\subsection{Public Health Evaluation Tasks and Datasets} \label{sec_tasks}

\begin{figure}[h]
    \centering
    \includegraphics[scale=0.39]{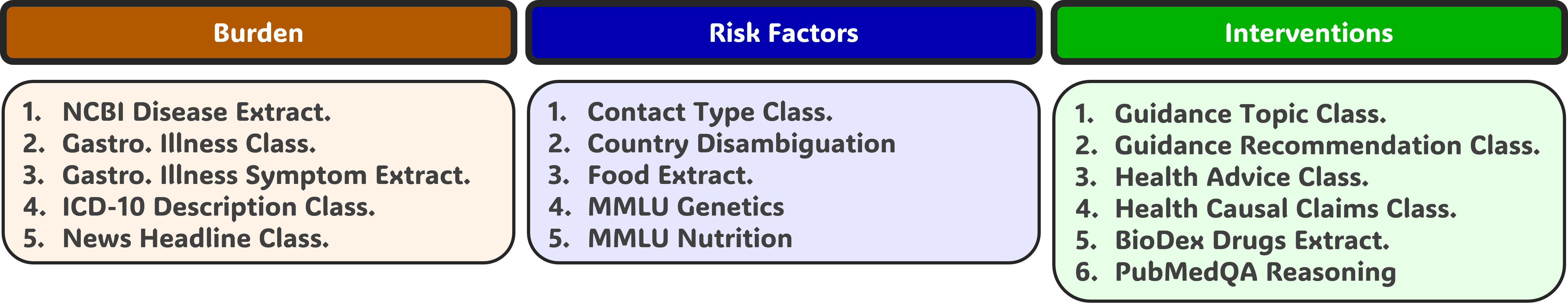} 
    \caption{\textbf{Evaluation Tasks by Public Health Area.} A summary of different tasks which we use to evaluate the LLMs, grouped by public health area. See \ref{Burden}, \ref{risk_factors}, and \ref{interventions} for full descriptions.}\label{fig:task_summary_fig}
\end{figure}

Public health is a broad field, intersecting with many different fields and issues in addition to conditions treated within healthcare settings \cite{winslow1920untilled, dahlgren1991policies, whitehead1992concepts, ross2015chapter}. Therefore, to ensure our evaluations are representative, the 16 classification and extraction tasks we bring together target a commensurately broad range of tasks within three key sub-domains of public health: burden (\ref{Burden}), risk factors (\ref{risk_factors}), and interventions (\ref{interventions}), as shown in \Cref{fig:task_summary_fig}. 

The range of data used in our evaluation is also broad, because potentially relevant health information is found in a diverse range of sources: academic literature, electronic health records, public health guidance, social media, news articles, and questionnaire responses ~\cite{vayena2018policy}. Details of all the tasks, datasets, and annotations are shown in \Cref{tab:datasets}.

\subsubsection{Burden}
\label{Burden}
Public health aims to mitigate adverse health outcomes in the population, which requires gathering information on health burden such as reports of symptoms, injuries, cases, morbidity, or mortality \cite{vos2020global}. Systematic data collection on burden is critical for developing evidence-based public health measures \cite{nsubuga2011public, who2016action, assembly2018political}. We use the following tasks to evaluate LLMs in this sub-domain: 

\begin{enumerate} [label=\bfseries\arabic*.,leftmargin=*,align=left]
    \item \textit{NCBI Disease Extraction:} To evaluate an LLM's ability to identify and extract diseases from free text, we use the NCBI disease corpus~\cite{dougan2014ncbi} of PubMed article abstracts annotated with the diseases mentioned and their associated MeSH (Medical Subject Headings) and OMIM (genes and genetic disorders)  codes. This task involves prompting the LLM to extract a structured comma separated list of diseases from the free text. In order to relieve some of the issues observed in the literature around exact matching of output strings~\citep{agrawal2022large, doosterlinck2023biodex, shyr2023identifying}, we first map all extracted disease mentions to their respective codes and assess performance on the de-duplicated set of MeSH and OMIM codes. 
    \item \textit{Gastrointestinal Illness Classification:}\label{gi_data} To evaluate an LLM's ability to identify potential illness or disease within non-technical social media free text, we use the Yelp Open Dataset~\cite{yelpopendataset} of restaurant reviews. To annotate the dataset, we first filter to those reviews that contain at least one of a comprehensive list of GI illness related keywords. We then take a random sample of approximately 3000 restaurant reviews and manually annotate (\Cref{annotations}) whether they refer to an instance of possible GI illness using an agreed epidemiological protocol.
    
    The LLM is prompted to provide a binary classification of GI illness ("yes" or "no") for each review. This is an adversarial task as all reviews manually annotated as "no" do contain at least one keyword associated with possible GI illness. 
    \item \textit{Gastrointestinal Illness Symptom Extraction:}\label{gi_data_subset} To evaluate an LLM's ability to extract possible symptoms from non-technical social media free text, we use the same annotated Yelp review dataset as in~\ref{gi_data} but filter to only those annotated as referring to possible GI illness. We then annotate (\Cref{annotations}) these reviews with all symptoms mentioned within the free text. The LLM is then prompted to extract all symptoms as a structured comma separated list.
    \item \textit{ICD-10 Description Classification:}\label{icd_10_data} In order to evaluate an LLM's ability to identify infections and conditions attributed to infections, we use descriptions of \textit{abnormal findings}, \textit{signs of illness} and \textit{symptoms} from the International Statistical Classification of Diseases and Related Health Problems (ICD)~\cite{who2016icd10} classification system. Using a protocol, two research analysts with relevant expertise separately annotate ICD-10 Version:2019 descriptions (\Cref{annotations}) with whether they directly refer to an infection or to a disease with a primarily infectious aetiology. We use a balanced sample of infection and non-infection disease descriptions. The LLM is then prompted to provide a binary classification of whether an ICD-10 code description relates to an infection, using the description and a summary of the classification protocol. 
    \item \textit{News Headline Classification:} We evaluate an LLM's ability to identify references to infectious diseases within non-technical free text using a manually annotated (\Cref{annotations}) dataset of news headlines with possible references to avian influenza collected from the GDELT Project~\cite{gdelt}. The LLM is prompted with a set of 5 news headlines and asked to provide a structured JSON output with its classifications. A secondary purpose of this task is to evaluate the LLM's ability to generate correctly formatted JSON strings consistently.
    \item \textit{\textbf{**Removed**} - MMLU Virology:} In our initial evaluations we used the MMLU Virology~\cite{mmlu} subset to assess an LLM's basic knowledge of virology. However, subsequent research into the error rate and quality of this subset~\cite{gema2024mmlu} means we no longer include it.
\end{enumerate}

\definecolor{Apricot}{rgb}{1.0, 0.68, 0.48}
\definecolor{SkyBlue}{rgb}{0.53, 0.81, 0.92}
\definecolor{Honeydew}{rgb}{0.94, 1, 0.94}

\begin{table*}[htbt!]
\centering
    \resizebox{\textwidth}{!}{
\begin{tabular}{@{}p{5cm}|p{3cm}|c|c|c|c|c|p{3cm}|c@{}}
\toprule
\multicolumn{1}{l|}{\multirow{2}{*}{\bf Task Name}} &
\multicolumn{1}{c|}{\multirow{2}{*}{\bf Dataset}} &
\multicolumn{1}{c|}{\multirow{2}{*}{\bf Task Type}} &
  \multicolumn{2}{c|}{\bf Test Set Size} &
  \multicolumn{1}{c|}{\multirow{2}{*}{\bf Text Type}} &
  \multicolumn{1}{c|}{\multirow{2}{*}{\bf Text Len}} &
  \multicolumn{1}{c|}{\multirow{2}{*}{\bf Example Labels}} &
  \multirow{2}{*}{\bf Public} \\ \cmidrule(lr){4-5}
\multicolumn{1}{l|}{} &
  \multicolumn{1}{c|}{} &
  \multicolumn{1}{c|}{} &
  \multicolumn{1}{c|}{\bf Rows} &
  \multicolumn{1}{c|}{\bf Labels} &
  \multicolumn{1}{c|}{} &
  \multicolumn{1}{c|}{ (Avg char)} &
 \\ \midrule  \rowcolor{Apricot!20}                     NCBI Disease Extraction &      NCBI Disease Corpus &     Extraction &            475 &              907 & Academic &                   1276 & ["non-hereditary (sporadic) breast cancer", "br... &                Yes \\
\midrule \rowcolor{Apricot!20}    Gastro-intestinal Illness Classification &        Yelp Open Dataset & Classification &           2456 &             2456 &       Social Media &                    635 &                                              "Yes" &                 No \\
\midrule \rowcolor{Apricot!20}  Gastro-intestinal Illness Symptom Extraction &        Yelp Open Dataset &     Extraction &            400 &              461 &       Social Media &                    464 &                                         ["nausea"] &                 No \\
\midrule     \rowcolor{Apricot!20}        ICD-10 Description Classification &                   ICD-10 & Classification &           2226 &             2226 &  Academic &                     36 &                                              "Yes" &                 No \\
\midrule  \rowcolor{Apricot!20}                News Headline Classification &                      GDELT & Classification &            353 &              353 &      News Articles &                     70 &                                              "Yes" &                 No \\
\midrule        \rowcolor{SkyBlue!20}           Contact Type Classification & Synthetic Questionnaires & Classification &            254 &              254 &      Questionnaire &                     86 &                                           "Rule 2" &                 No \\
\midrule       \rowcolor{SkyBlue!20}                Country Disambiguation &    GP Registration Forms & Classification &           8000 &             8000 &      Questionnaire &                     14 &                                           "Brazil" &                 No \\
\midrule       \rowcolor{SkyBlue!20}                       Food Extraction &        Yelp Open Dataset &     Extraction &            400 &              602 &       Social Media &                    464 &                                  ["fish", "fruit"] &                 No \\
\midrule      \rowcolor{SkyBlue!20}                         MMLU Genetics  &                     MMLU & Classification &             93 &               93 &    Multiple Choice &                      - &                                                "C" &                Yes \\
\midrule         \rowcolor{SkyBlue!20}                     MMLU Nutrition  &                     MMLU & Classification &            276 &              276 &    Multiple Choice &                      - &                                                "C" &                Yes \\
\midrule      \rowcolor{Honeydew!40}          Guidance Topic Classification &                    UKHSA & Classification &            265 &              265 &           Guidance &                    456 &                                                "4" &                 No \\
\midrule  \rowcolor{Honeydew!40}     Guidance Recommendation Classification &                    UKHSA & Classification &            392 &              392 &           Guidance &                    794 &                                              "Yes" &                 No \\
\midrule     \rowcolor{Honeydew!40}            Health Advice Classification &             HealthAdvice & Classification &           8676 &             8676 &           Academic &                    144 &                                                "2" &                Yes \\
\midrule  \rowcolor{Honeydew!40}        Health Causal Claims Classification &           CausalRelation & Classification &           2448 &             2448 &           Academic &                    125 &                                                "1" &                Yes \\
\midrule    \rowcolor{Honeydew!40}                                 PubMedQA &                 PubMedQA & Classification &            800 &              800 &           Academic &                   1330 &                                              "Yes" &                Yes \\
\midrule     \rowcolor{Honeydew!40}                 BioDex Drugs Extraction &                   BioDex &     Extraction &           1247 &             4429 &  Academic &                 c.6000 &                    ["flucloxacillin", "midazolam"] &                Yes \\
 \bottomrule
\end{tabular}}
\caption{\textbf{Overview of Public Health Evaluation Tasks}. In order to capture a broad range of free text, the 16 tasks we use draw on 13 distinct datasets from internal and external sources. "Text length" refers to the average number of characters in the free text (excluding the prompt template and question). "Public" refers to whether the annotations are available online.}
\label{tab:datasets}
\end{table*}

\subsubsection{Risk Factors}\label{risk_factors}
Epidemiological risk factors are environmental, behavioral, or biological factors that increase a person's likelihood of developing disease or injury \cite{murray2020global}. Risk factor surveillance is essential for quantifying these risks and developing evidence-based interventions to reduce them \cite{mcgeehin2004national}. Addressing risk factors directly may provide a more effective strategy than treating diseases once they arise \cite{who2013global, martin2020ounce}. We use the following tasks to evaluate LLMs in this sub-domain:

\begin{enumerate} [label=\bfseries\arabic*.,leftmargin=*,align=left]
    \item \textit{Contact Type Classification:} A key challenge during outbreak and pandemic response is often rapidly implementing and scaling contact tracing~\cite{cmo_cov}. One important aspect of this is identifying the type of contact that has occurred in order to assess the risk of onward transmission. To evaluate an LLM's ability to identify contact types from representative free text, we generate and manually annotate (\Cref{annotations}) an entirely new synthetic dataset created using GPT-4 via the OpenAI API~\cite{openaiapi}, designed to reflect the style, content and structure of the answers provided within the enhanced surveillance questionnaires for contacts submitted during the mpox outbreak response~\cite{mpox_ct}. We prompt the LLM to classify the type of contact based on an epidemiological protocol.
    
    This task is challenging for two reasons. First, it requires the LLM to apply a detailed protocol provided within the prompt, rather than drawing on existing knowledge provided during pre-training. Second, this particular dataset was chosen because the text often contains discussion of sexual activity, which is an important risk factor for certain infections~\cite{sonnenberg2013prevalence}. However, many LLM pre-training~\cite{raffel2023exploring} and fine-tuning datasets~\cite{touvron2023llama} are designed to avoid text about sexual activity and so evaluating performance on this type of free text is essential if using LLMs for certain disease areas in public health.     
    \item \textit{Country Disambiguation:} Different pathogens are endemic to different regions of the world~\cite{murray2020global}. As such, understanding the risk profile of an individual often requires understanding their recent travel history or previous countries where they have lived. Identifying geographies within free text is often an important task to help determine an individual's risk of infection or other exposure~\cite{ng2020application}. To evaluate an LLM's knowledge and understanding of geographic locations, we use a manually annotated dataset (\Cref{annotations}) of anonymised free text responses from GP registration form place of birth fields where the location cannot be identified using existing automated matching. The main reasons matches fail is people supply place names within countries (without reporting the country) and typographical errors. The LLM is prompted to either disambiguate the country the free text refers to or identify it as unknown. The LLM response is then post-processed to a standardised list of countries using the \textit{country-converter} Python package ~\cite{Stadler2017}.       
    \item \textit{Food Extraction:} To evaluate an LLM's ability to generate structured data on risk factors from social media free text, we use the same filtered annotated Yelp review dataset as in \textit{Gastrointestinal Illness Symptom Extraction} (\ref{gi_data_subset}). We then manually annotate these reviews with all the foods mentioned within the free text. The LLM is prompted to extract all references to food or meals as a structured comma separated list. Foods are very challenging to disambiguate, so we use a large lookup table of foods based on the FoodEx 2 database \cite{european2015food} to disambiguate the foods the LLM extracts into a list of 27 potential labels that are relevant to public health foodborne illness investigation. 
    \item \textit{MMLU Genetics:} To evaluate an LLM's basic knowledge of genetics, we use the Medical Genetics subset of the MMLU benchmark~\cite{mmlu} (\Cref{sub_set_lit}). The LLM is prompted to provide the answer to multiple choice questions on a range of topics within medical genetics.  
    \item \textit{MMLU Nutrition:} Similarly, to evaluate an LLM's basic knowledge of nutrition, we use the Nutrition subset of the MMLU benchmark~\cite{mmlu} (\Cref{sub_set_lit}). The LLM is prompted to provide the answer to multiple choice questions on a range of topics within nutrition.  
\end{enumerate}

\subsubsection{Interventions}\label{interventions}
Public health interventions can take many forms ~\cite{ross2015chapter} ~\cite{fortune2021international}. One common way to classify them is into pharmaceutical and non-pharmaceutical interventions ~\cite{mendez2021systematic}. Pharmaceutical interventions predominantly use medical technologies to prevent, diagnose or treat disease. Vaccination is one particularly common public health pharmaceutical intervention. Public health non-pharmaceutical interventions generally aim to reduce behavioural risk factors and so mitigate various forms of disease. These include nutrition advice or infection control through hand hygiene.

Public health guidance is one of the primary ways public health interventions can be communicated and implemented. Therefore, we introduce four initial guidance related evaluations of increasing difficulty to assess LLMs. We also evaluate LLMs for processing text related to pharmaceutical interventions and biomedical reasoning. 
\begin{enumerate} 
[label=\bfseries\arabic*.,leftmargin=*,align=left]
    \item \textit{Guidance Topic Classification:} To evaluate an LLM’s ability to identify the relevant topics to which a piece of guidance relates, we used 331 UKHSA guidance publication summary pages from the gov.uk web page. The text was extracted and manually categorised (\Cref{annotations}) into one of eight broad health topics based on which team wrote the guidance (e.g. Radiation, Sexually Transmitted Infections, etc.). The LLM is prompted with the summary page and the list of possible health topics and asked to return the single health topic that is most suitable for the text.
    \item \textit{Guidance Recommendation Classification:} \label{guidance_rec} Recommendations are a crucial component of public health guidance. Demonstrating an LLM's ability to identify recommendations is an essential prerequisite to other processing of public health guidance. To evaluate this we use UKHSA publications on gov.uk. Each publication is split into subsections (chunks) and we manually annotate a random sample of 489 chunks of text, with one of two labels: 1 (contains recommendations – defined as a statement containing a clear, specific actionable instruction, request, or advice in the event of a given public health scenario), or 0 (does not contain any recommendations). The LLM is prompted with the text chunk and asked to answer "yes" or "no" (corresponding to class 1 or 0) to whether it contains any recommendations.
    \item \textit{Health Advice Classification:} To further evaluate an LLM’s ability to understand recommendations, we adopt the HealthAdvice dataset from \citet{Chen_2024} (\Cref{drug_int_lit}), which includes 10,845 manually annotated sentences from the abstract and discussion sections of PubMed articles. Each sentence has one of three labels: 0 (no advice), 1 (weak advice – statement hints that a behaviour or practice may require changing, or that an alternative approach for existing clinical or medical practice may be required), 2 (strong advice – statement makes a clear and straightforward recommendation for a change in behaviour or practice)~\cite{yu2019EMNLPCausalLanguage}. The LLM is prompted with the sentence and asked to assign the sentence a label of 0, 1, or 2, corresponding to the levels of health advice described. This extends~\ref{guidance_rec} to a different corpus (biomedical literature as opposed to public health guidance), shorter text (sentences rather than chunks), and advice slightly differing in definition from recommendation.
    \item \textit{Health Causal Claims Classification:} The final guidance-related task we consider is evaluating an LLM’s understanding of the types of claims that can be found within guidance and broader biomedical text. To do this, we use the Causal-Relation dataset~\cite{Chen_2024} (\Cref{drug_int_lit}), which consists of annotated biomedical text from PubMed article conclusions. Each sentence is labelled as one of the following classes: 0 (no relationship), 1 (correlational relationship – association between variables are described, but causation is not explicitly stated), 2 (conditional causal relationship – suggestion that one variable directly changes the other, with an element of doubt), or 3 (direct causal – explicit statement that one variable directly changes the other)~\cite{li2021EMNLPHealthAdvice}. The LLM is prompted with the sentence and asked to assign a label of 0, 1, 2, or 3, based on the relationship described.
    \item \textit{BioDex Drugs Extraction:} To evaluate an LLM's ability to extract information on pharmaceutical interventions we use 1,558 randomly sampled entries from the  BioDex dataset~\cite{doosterlinck2023biodex} (\Cref{drug_int_lit}) of annotated Adverse Drug Event reports and corresponding PubMed articles. To account for different model context windows we first chunk each article into sets of contiguous paragraphs of no more than 6000 characters. The LLM is prompted to extracted the drugs involved in the ADE from each of the raw PubMed article free text chunks. 
    \item \textit{PubMedQA:} Text regarding guidance and interventions often includes supporting evidence and relevant academic findings. Therefore, we also evaluate an LLM's ability to understand and reason about biomedical evidence. To evaluate an LLM on this task we use the PubMedQA~\cite{jin2019pubmedqa} benchmark in the "reasoning-required setting". The LLM is prompted with a PubMed abstract (excluding any concluding sections) and asked to answer the question ("yes", "no", or "maybe") that is posed in the title of the article. The LLM's performance is evaluated against expert human annotator answers.  
\end{enumerate}

\subsubsection{Summary}

In constructing these evaluations we have balanced general public health text processing (such as drugs and disease extraction) with specific public health tasks (such as contact and guidance classification). We also combine external evaluation datasets for comparability and diversity of text with internal datasets that are more representative of public health specific tasks and free text. 

\subsection{Evaluation Methodology}

Our aim is to provide a consistent assessment of LLMs across tasks, rather than attempt to achieve the highest possible performance for any given model and task combination. Therefore, in this paper we focus on implementing simple but standardised prompting, post-processing, and sampling across all models. As discussed in \Cref{sec:results}, this means our zero-shot prompt\footnote{The number of "shots" refers to how many example question-answer pairs are provided in the prompt, in addition to the question being asked.} evaluations should provide a reasonable lower bound for each model's performance, with potentially significant gains possible from more advanced pipelines.

\begin{table*}[htbt!]
\centering
    \resizebox{\textwidth}{!}{
\begin{tabular}{@{}p{5cm}|c|c|c|c|c|c|c@{}}
\toprule
\multicolumn{1}{l|}{\multirow{2}{*}{\bf Model Name}} &
\multicolumn{1}{c|}{\multirow{2}{*}{\bf Base Model}} &
\multicolumn{1}{c|}{\multirow{2}{*}{\bf Precision}} &
  \multicolumn{2}{c|}{\bf Model Size} &
  \multicolumn{1}{c|}{\multirow{2}{*}{\bf Author}} &
  \multirow{2}{*}{\bf Host} \\ \cmidrule(lr){4-5}
\multicolumn{1}{l|}{} &
  \multicolumn{1}{c|}{} &
  \multicolumn{1}{c|}{} &
  \multicolumn{1}{c|}{\bf Params (bn)} &
  \multicolumn{1}{c|}{\bf Context (toks)} &
  \multicolumn{1}{c|}{} &
 \\  \midrule Flan-T5-xxl~\cite{flan} & Flan-T5-xxl & FP16 & 11 & 2048 & Google & Internal API \\
Mistral-7B-Instruct-v0.2~\cite{jiang2023mistral} & Mistral-7B & FP16 & 7 & 32768 & Mistral & Internal API \\
Mistral-Large-Instruct & Mistral-Large & INT4-AWQ & 123 & 128000 & Mistral & Internal API \\
Stable-Beluga-2~\cite{StableBelugaModels} & Llama-2-70B~\cite{touvron2023llama} & FP16 & 70 & 4096 & Stability-AI + Meta & Internal API \\
Llama-3-70B-Instruct~\cite{llama3modelcard} & Llama-3-70B & FP16 & 70 & 8192 & Meta & Internal API \\
Llama-3-8B-Instruct~\cite{llama3modelcard} & Llama-3-8B & FP16 & 8 & 8192 & Meta & Internal API \\
Llama-3-1-70B-Instruct~\cite{llama3_1} & Llama-3-1-70B & FP16 & 70 & 128000 & Meta & Internal API \\
Llama-3-1-70B-Instruct~\cite{llama3_1} & Llama-3-1-70B & INT4-AWQ & 70 & 128000 & Meta & Internal API \\
Llama-3-1-8B-Instruct~\cite{llama3_1} & Llama-3-1-8B & FP16 & 8 & 128000 & Meta & Internal API \\
Llama-3-1-8B-Instruct~\cite{llama3_1} & Llama-3-1-8B & INT4-AWQ & 8 & 128000 & Meta & Internal API \\
Llama-3-3-70B-Instruct~\cite{llama3_3} & Llama-3-1-70B & FP16 & 70 & 128000 & Meta & Internal API \\
Llama-3-3-70B-Instruct~\cite{llama3_3} & Llama-3-1-70B & INT4-AWQ & 70 & 128000 & Meta & Internal API \\
Gemma-2-27b-it~\cite{gemma2} & Gemma-2-27b & FP16 & 27 & 8192 & Google & Internal API \\
Command-R~\cite{commandr} & Command-R & FP16 & 32 & 128000 & Cohere & Internal API \\
gpt-4-turbo-2024-04-09 & GPT-4 & - & - & 128000 & OpenAI & OpenAI API \\
gpt-4o-mini-2024-07-18 & GPT-4o-mini & - & - & 128000 & OpenAI & OpenAI API \\
gpt-4o-2024-11-20 & GPT-4o & - & - & 128000 & OpenAI & OpenAI API \\
\bottomrule
\end{tabular}}
\caption{\textbf{Overview of Large Language Models Evaluated.} We focus on evaluating open-source/weight LLMs that we host internally to ensure comparability of results, as well as for data protection. We run all models in FP16 precision where feasible. We include the latest GPT-4 models for comparison on a subset of tasks.}
\label{tab:models}
\end{table*}

\subsubsection{Large Language Models}

In our initial evaluations we assess eleven open-weight LLMs ranging from 7 billion to 123 billion parameters (including the Llama-2 and 3, Mistral, Command-R, Gemma, and Flan-T5 base models), see \Cref{tab:models}. We also investigate the performance impact of INT-4 quantization across the Llama-3 family of models. 

To run the evaluations, we have developed an internal LLM API on UKHSA High Performance Computing (HPC) resources~\cite{ukhsa_ai_board} using models from the HuggingFace repository~\cite{HuggingFaceGeneral} and open-source packages such as \textit{transformers}~\cite{wolf2020huggingfaces} and \textit{vLLM}~\cite{kwon2023efficient}. This enables us to securely use datasets where data governance and security mean they are not authorised to leave UKHSA systems. It also allows us to control implementation details, such as model quantisation, prompt templates, model versions, and generation configurations, that have been shown to have potentially significant impacts on performance~\cite{huang2024good}, allowing for comparable and reproducible results.

In addition to evaluating internally hosted models, we also access some of the highest performing private models (GPT-4 Turbo and GPT-4o), via the OpenAI API~\cite{openaiapi}. While we do not run GPT-4 series models for all evaluations primarily due to data restrictions, we provide results for GPT-4-Turbo, GPT-4o, and GPT-4o-Mini on a subset of 11 tasks across the three areas for comparison. 

\subsubsection{Prompting}
For all tasks, we report zero-shot prompting results. Additionally, for a subset of more complex tasks we also investigate few-shot prompting results. 

We use the prompt templates provided by the model authors where available. This leads to some variation in prompt content due to the availability of system prompts in some templates. However, we aim to keep the informational content consistent across prompts as far as possible.

\subsubsection{Sampling}
For all tasks, we use greedy decoding in order to generate reproducible results. 

\subsubsection{Dataset Splits}
We use a 20-80 validation-test split for all tasks, where only the validation set is used for prompt development; this avoids overfitting the prompt to the test set, which we use to quantify performance in this paper. 

\subsubsection{Evaluation Metrics}

In this paper we primarily report the micro and macro F1 scores for each task. The F1 score is calculated as the harmonic mean of precision\footnote{True Positives divided by the total number of True Positives and False Positives} (positive predictive value) and recall\footnote{True Positives divided by the total number of True Positives and False Negatives} (sensitivity). The micro-F1 score is calculated by weighting each label according to its frequency in the dataset, while the macro-F1 score is calculated weighting each label equally. It is also important to note that for single label classification tasks the micro-F1 score is equivalent to accuracy.\label{f1_def} 

Our headline measure of performance is micro-F1, as we are assessing raw performance rather than accounting for potentially heterogeneous public health significance of different labels.

For extraction tasks, we use exact matching of output strings. We treat extracted outputs not found in the ground-truth label set as valid (i.e included in result calculations) but incorrect classifications (because they are not one of the ground-truth label options).

For classification tasks, we also use exact matching but with simple post-processing to clean outputs (e.g we would convert "Rule 1" -> "rule 1", and "1." -> 1). 

However, for some tasks with large numbers of possible labels and where it is feasible, we implement more advanced post-processing to standardise outputs, such as for \textit{Country Disambiguation}, \textit{NCBI Disease Extraction}, and \textit{Food Extraction}.

\section{Results} \label{sec:results}
We primarily discuss results for internally hosted open-weight models as they can be evaluated for all tasks. We compare these results with GPT-4 series model performance on a subset of tasks in \Cref{gpt_results}. 

\definecolor{Apricot}{rgb}{1.0, 0.68, 0.48}
\definecolor{SkyBlue}{rgb}{0.53, 0.81, 0.92}
\definecolor{Honeydew}{rgb}{0.94, 1, 0.94}
\definecolor{Grey}{rgb}{0.5, 0.5, 0.5}

\begin{table*}[htbt!]
\centering
    \resizebox{\textwidth}{!}{
\begin{tabular}{lccccccccccc}
\toprule
 & \makecell{Mistral\\7B} & \makecell{Llama-3\\8B} & \makecell{Llama-3.1\\8B} & \makecell{Flan-T5\\XXL} & \makecell{Gemma-2\\27B-It} & \makecell{Command\\R} & \makecell{Stable\\Beluga-2} & \makecell{Llama-3\\70B} & \makecell{Llama-3.1\\70B} & \makecell{Llama-3.3\\70B} & \makecell{Mistral\\Large*} \\
Task Name &  &  &  &  &  &  &  &  &  &  &  \\
\midrule
\rowcolor{Apricot!20} NCBI Disease Extraction & 0.63 & 0.80 & 0.87 & 0.72 & 0.83 & 0.83 & 0.82 & 0.84 & 0.83 & \underline{\textbf{0.88}} & 0.83 \\
\rowcolor{Apricot!20} GI Classification & 0.86 & 0.82 & 0.84 & 0.89 & 0.82 & 0.91 & 0.90 & 0.92 & \underline{\textbf{0.94}} &  \underline{\textbf{0.94}} & \underline{\textbf{0.94}} \\
\rowcolor{Apricot!20} GI Symptom Extraction & 0.58 & 0.87 & 0.71 & 0.76 & 0.90 & 0.87 & 0.90 & \underline{\textbf{0.91}} & 0.90 & 0.90 & 0.90 \\
\rowcolor{Apricot!20} ICD-10 Description Classification & 0.94 & 0.93 & 0.85 & 0.92 & 0.94 & 0.87 & 0.94 & \underline{\textbf{0.95}} & 0.94 & 0.94 & 0.94 \\
\rowcolor{Apricot!20} News Headline Classification & 0.90 & 0.95 & 0.95 & 0.44 & 0.95 & \underline{\textbf{0.97}} & 0.95 & 0.93 & 0.93 & 0.94 & 0.90 \\
\rowcolor{Honeydew!40} Guidance Topic Classification & 0.73 & 0.87 & 0.82 & 0.94 & 0.91 & 0.87 & 0.91 & 0.94 & 0.93 & \underline{\textbf{0.96}} & 0.89 \\
\rowcolor{Honeydew!40} BioDex Drugs Extraction & 0.25 & 0.32 & 0.32 & 0.33 & \underline{\textbf{0.36}} & 0.30 & 0.30 & 0.33 & 0.35 & 0.34 & 0.33 \\
\rowcolor{Honeydew!40} Health Advice & 0.75 & 0.68 & 0.78 & 0.77 & 0.74 & 0.78 & 0.66 & 0.78 & 0.81 & \textbf{0.82} & 0.82 \\
\rowcolor{Honeydew!40} Guidance Recommendations & 0.75 & 0.83 & 0.86 & 0.83 & 0.82 & \underline{\textbf{0.88}} & 0.82 & 0.87 & 0.86 & 0.86 & 0.87 \\
\rowcolor{Honeydew!40} Causal Relations & 0.38 & 0.26 & 0.39 & 0.50 & 0.45 & 0.42 & 0.35 & 0.51 & 0.53 & \textbf{0.62} & 0.58 \\
\rowcolor{Honeydew!40} PubMedQA & 0.33 & 0.74 & 0.76 & 0.76 & 0.49 & 0.73 & 0.73 & 0.75 &  \underline{\textbf{0.77}} & \underline{\textbf{0.77}} & 0.59 \\
\rowcolor{SkyBlue!20} Contact Classification & 0.24 & 0.23 & 0.26 & 0.28 & 0.42 & 0.36 & 0.46 & 0.48 & 0.49 & \underline{\textbf{0.53}} & 0.22 \\
\rowcolor{SkyBlue!20} Country Disambiguation & 0.82 & 0.86 & 0.89 & 0.83 & 0.91 & 0.92 & 0.86 & 0.92 & \underline{\textbf{0.93}} & 0.92 & 0.90 \\
\rowcolor{SkyBlue!20} Food Extraction & 0.87 & 0.87 & 0.88 & 0.83 & 0.88 & 0.88 & 0.88 & 0.88 & 0.88 & \textbf{0.89} & 0.87 \\
\rowcolor{SkyBlue!20} MMLU Genetics & 0.59 & 0.70 & 0.72 & 0.59 & 0.90 & 0.78 & 0.68 & \textbf{0.91} & \textbf{0.91} & 0.86 & 0.90 \\
\rowcolor{SkyBlue!20} MMLU Nutrition & 0.60 & 0.68 & 0.72 & 0.60 & 0.81 & 0.74 & 0.71 & 0.85 & \textbf{0.88} & 0.86 & \textbf{0.88} \\
\midrule
\rowcolor{Grey!10} Mean Task Rank & 9.50 & 8.44 & 6.56 & 7.75 & 5.75 & 5.88 & 7.31 & 3.12 & 2.94 & \textbf{2.50} & 5.31 \\
\bottomrule
\end{tabular}
}
\caption{\textbf{Zero-shot Results (Micro-F1 Scores) for Open-Weight Models.} Bold indicates the highest open-weight model micro-F1 score on a given task. Underlined indicates the highest micro-F1 score across all models evaluated (inc. GPT-4 series for a subset of tasks). Mean Task Rank is calculated as the rank of the open-weight model on each task averaged over all tasks. Models ordered by number of parameters. *Mistral-Large was run in INT-4 AWQ}
\label{tab:results}
\end{table*}

\subsection{Model Results}
\Cref{tab:results} shows the performance of models on different tasks. Llama-3.3-70B-Instruct is the highest scoring open-weight model (or equal highest) across 8 of the 16 tasks according to micro-F1. 
 
Flan-T5-xxl, Llama-3-8B-Instruct, Llama-3.1-8B-Instruct, and Mistral-7B-Instruct-v0.2 were each the worst performing model on at least one task. The authors note that smaller models appear to have outputs that are more fragile with respect to the exact wording of the prompt. This means that particularly low performance on a given task may be more related to a failure of the model to understand the prompter's intentions, rather than the model not being capable of performing the task with optimised prompts. 

On some tasks the difference in performance between models is very considerable. For instance, scaling parameter size from the Llama-3-8B-Instruct to the Llama-3-70B-Instruct leads to greater than 10 percentage point increases in micro-F1 on tasks such as MMLU Genetics, MMLU Nutrition, Health Advice, and Causal Relation classification (smaller gains are observed for the 3.1 series models). Similarly, moving from a Llama-2-70B base model to a Llama-3-70B base model also considerably improves performance on the same tasks. 

We observe incremental gains across the Llama 3 70bn model family with each version having a higher mean task rank than the previous. We also note that Mistral-Large's relatively poor performance for its size may be due to our implementation using INT-4 quantization vs FP16 for the other models, due to compute constraints.  

\subsection{Task Results}
We find significant variability in performance across tasks. For tasks like \textit{ICD-10 Description Classification} all of the models classify text with a high level of accuracy, with the minimum micro-F1 score achieved 85\%. Similarly, for other tasks like \textit{Gastrointestinal Illness Classification}, \textit{Country Disambiguation}, and \textit{Food Extraction}, all models achieve micro-F1 scores above 80\%.

However, some tasks still appear challenging for the LLMs evaluated, with no LLM getting over 60\% micro-F1 on any of \textit{BioDex Drugs Extraction}, and \textit{Contact Classification} using zero-shot prompts.

\subsection{Few-shot Prompting}

To investigate the potential impact of using more advanced prompting techniques on challenging tasks, we evaluate the impact of few-shot prompting on the hardest internal and external classification tasks.

\begin{figure}[ht]
    \centering
    \includegraphics[width=\textwidth]{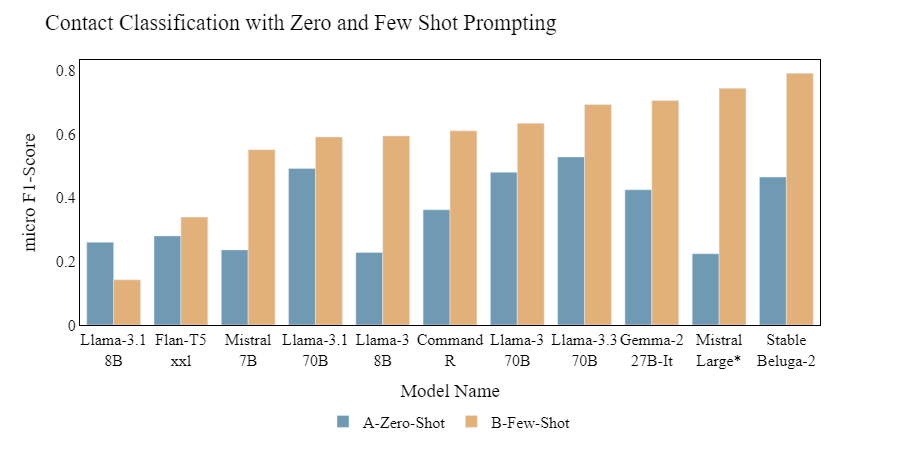}
    \caption{\textbf{Comparison of zero-shot and few-shot prompting on challenging tasks (Contact Classification).} We compare the baseline zero-shot prompt to a 10-shot prompt for \textit{Contact Classification}.}
    \label{fig:contact_few_shot}
\end{figure}

For the hardest of our internally annotated tasks, \textit{Contact Classification}, we use a 10-shot prompt illustrating how to apply the protocol in difficult edge cases. We find substantial improvements across models, see \Cref{fig:contact_few_shot}. With the exception of Flan-T5-xxl and Llama-3.1-8B, all other models see greater than 10 percentage point increases in micro-F1 scores over the zero-shot setting.

\begin{figure}[ht]
    \centering
    \includegraphics[width=\textwidth]{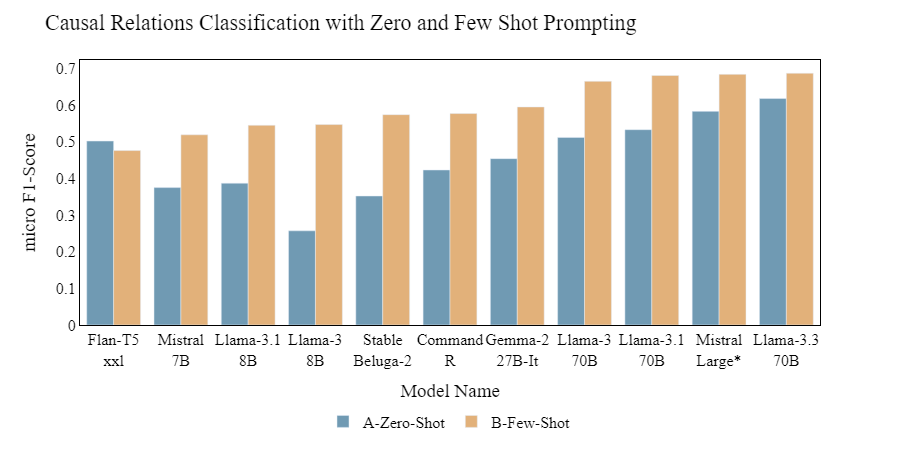}
    \caption{\textbf{Comparison of zero-shot and few-shot prompting on challenging tasks (Health Causal Claims Classification).} We compare the baseline zero-shot prompt to a 7-shot prompt for \textit{Health Causal Claims Classification}.}
    \label{fig:causal_few_shot}
\end{figure}

\begin{figure}[H]
    \centering
    \includegraphics[scale=0.6]{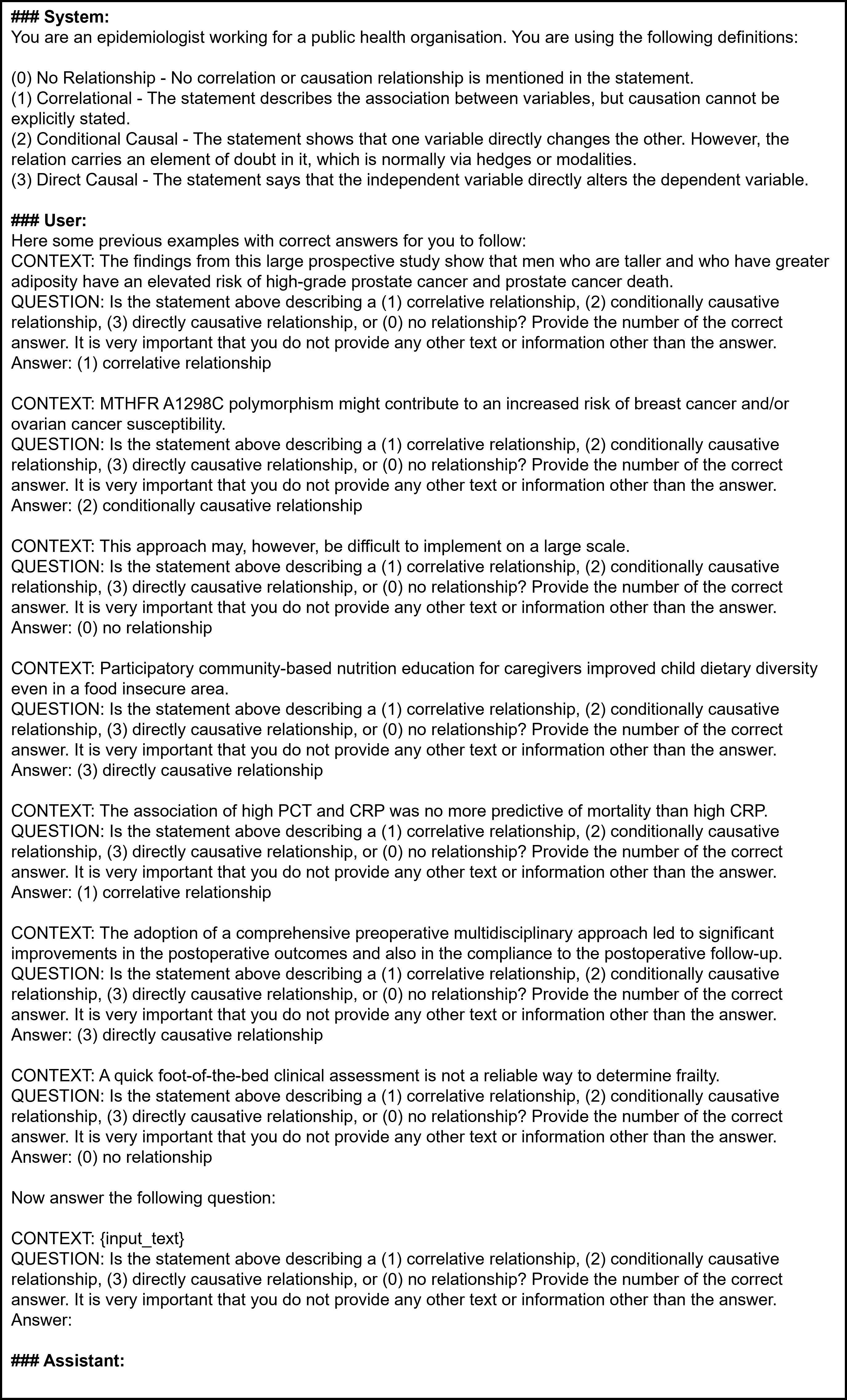}
    \caption{\textbf{Example Few Shot Prompt for Health Causal Claims Classification}. This is an example of the 7 shot prompt used in the \textit{Health Causal Claims Classification} task structured with the Stable-Beluga-2 prompt template. Few-shot examples taken from \citet{yu2019EMNLPCausalLanguage}.}\label{fig:prompt_few_causal}
\end{figure}

Similarly, for the hardest of the externally annotated classification tasks, \textit{Health Causal Claims Classification}, we use a 7-shot prompt demonstrating the application of the definitions to example sentences (see \Cref{fig:prompt_few_causal}). Again with the notable exception of Flan-T5-xxl (which performs strongly in the zero-shot setting), we also see a generally greater than 10 percentage point increase in micro-F1 scores over the zero-shot baseline across models, see \Cref{fig:causal_few_shot}. Large improvements in performance from few-shot prompting on this task were also found for GPT-4 and GPT-3.5 in the original work by \citet{Chen_2024}. 

The relative performance of models also changes between the zero-shot and few-shot settings for the two tasks. For example, Llama-3-8B-Instruct is the lowest performing model with a zero-shot prompt but outperforms the Flan-T5-xxl and Mistral-7B-Instruct-v0.2 models when using few-shot prompting. 

An interesting feature of both of these classification tasks, and potentially why there are significant gains from few-shot prompting, is that they require the LLM to make nuanced distinctions between complex labels that are defined by the user within the prompt. This type of task is particularly relevant for public health given the regular use of specific definitions and protocols that often change over time.

\subsection{GPT-4 and GPT-4o Comparison}\label{gpt_results}
\definecolor{Apricot}{rgb}{1.0, 0.68, 0.48}
\definecolor{SkyBlue}{rgb}{0.53, 0.81, 0.92}
\definecolor{Honeydew}{rgb}{0.94, 1, 0.94}
\definecolor{Grey}{rgb}{0.5, 0.5, 0.5}

\begin{table*}[htbt!]
\centering
\scriptsize
\begin{tabular}{lc|ccc}
\toprule
 & \makecell{Llama-3.3\\70B} & \makecell{GPT-4o\\Mini} & \makecell{GPT-4\\Turbo} & \makecell{GPT-4o} \\
Task Name &  &  &  &  \\
\midrule
\rowcolor{Apricot!20} NCBI Disease Extraction & \underline{\textbf{0.88}} & 0.79 & 0.83 & 0.85 \\
\rowcolor{Apricot!20} GI Classification & \underline{\textbf{0.94}} & 0.92 & 0.84 & 0.87 \\
\rowcolor{Apricot!20} GI Symptom Extraction & 0.90 & 0.90 & 0.89 & 0.86 \\
\rowcolor{Honeydew!40} BioDex Drugs Extraction & 0.34 & 0.33 & 0.33 & 0.34 \\
\rowcolor{Honeydew!40} Health Advice & \textbf{0.82} & 0.81 & 0.83 & \underline{0.85} \\
\rowcolor{Honeydew!40} Guidance Recommendations & 0.86 & 0.86 & 0.85 & 0.87 \\
\rowcolor{Honeydew!40} Causal Relations & \textbf{0.62} & 0.60 & \underline{0.64} & 0.57 \\
\rowcolor{Honeydew!40} PubMedQA & \underline{\textbf{0.77}} & 0.67 & 0.75 & 0.54 \\
\rowcolor{SkyBlue!20} Food Extraction & \textbf{0.89} & \underline{0.91} & 0.89 & 0.91 \\
\rowcolor{SkyBlue!20} MMLU Genetics & 0.86 & 0.91 & 0.96 & \underline{0.98} \\
\rowcolor{SkyBlue!20} MMLU Nutrition & 0.86 & 0.80 & 0.88 & \underline{0.91} \\
\bottomrule
\end{tabular}

\caption{\textbf{Zero-shot GPT-4 Results.} Bold indicates the highest open-weight model micro-F1 score on a given task. Underlined indicates the highest micro-F1 score across all models.}
\label{tab:gpt_4_results}
\end{table*}

Finally, to understand how current open-weight models compare on extraction and classification tasks to some of the highest performing private models, we evaluate three GPT-4 and GPT-4o models on 11 of the 16 tasks, see Table \ref{tab:gpt_4_results}.

We find GPT-4-Turbo, GPT-4o, and GPT-4o-Mini perform well across all tasks and are  the highest scoring models overall on 5 of the 11. The GPT-4 series models particularly outperforms open-weight models on \textit{MMLU Genetics}. However, looking across all tasks, the best GPT-4 models perform comparably to the most recent Llama-3.3-70B-Instruct open-weight model.

\subsection{Quantization Performance Impact}

The significant memory requirements of LLMs means a range of model quantization methods have been explored to reduce the total footprint of a model with a given number of parameters. These approaches change the format and precision in which the model parameters are stored to reduce the number of bits required for each parameter. However, the trade-off is that quantization will also impact model behavior, potentially degrading performance when compared to the model in its original format. 

\definecolor{Apricot}{rgb}{1.0, 0.68, 0.48}
\definecolor{SkyBlue}{rgb}{0.53, 0.81, 0.92}
\definecolor{Honeydew}{rgb}{0.94, 1, 0.94}
\definecolor{Grey}{rgb}{0.5, 0.5, 0.5}

\begin{table*}[htbt!]
\centering
\scriptsize
\begin{tabular}{lccc}
\toprule
 & \textbf{Llama-3.1-8B diff (ppts)} & \textbf{Llama-3.1-70B diff (ppts)} & \textbf{Llama-3.3-70B diff (ppts)} \\
\textbf{Task Name} &  &  &  \\
\midrule
\rowcolor{Apricot!20} NCBI Disease Extraction & 1.4 & -1.2 & 1.0 \\
\rowcolor{Apricot!20} GI Classification & 2.5 & 0.0 & 0.0 \\
\rowcolor{Apricot!20} GI Symptom Extraction & 0.0 & -0.8 & -0.4 \\
\rowcolor{Apricot!20} ICD-10 Description Classification & -4.1 & 0.2 & -0.2 \\
\rowcolor{Apricot!20} News Headline Classification & -0.9 & -0.2 & -0.5 \\
\rowcolor{Honeydew!40} Guidance Topic Classification & -4.2 & 1.5 & 0.3 \\
\rowcolor{Honeydew!40} BioDex Drugs Extraction & -0.4 & 0.6 & 0.3 \\
\rowcolor{Honeydew!40} Health Advice & -0.6 & -1.5 & -0.5 \\
\rowcolor{Honeydew!40} Guidance Recommendations & 2.0 & 0.0 & -1.3 \\
\rowcolor{Honeydew!40} Causal Relations & -8.8 & -1.7 & -2.4 \\
\rowcolor{Honeydew!40} PubMedQA & 0.6 & 0.0 & 0.1 \\
\rowcolor{SkyBlue!20} Contact Classification & -4.3 & 0.8 & -1.6 \\
\rowcolor{SkyBlue!20} Country Disambiguation & -1.3 & 0.1 & 0.0 \\
\rowcolor{SkyBlue!20} Food Extraction & 0.8 & -0.1 & 0.4 \\
\rowcolor{SkyBlue!20} MMLU Genetics & -2.1 & -1.1 & 2.2 \\
\rowcolor{SkyBlue!20} MMLU Nutrition & -1.5 & -0.3 & 0.0 \\
\midrule
\rowcolor{Grey!10}\textbf{Avg INT-4 AWQ vs FP16 Diff} & \textbf{-1.3} & \textbf{-0.2} &   \textbf{-0.2} \\
\bottomrule
\end{tabular}

\caption{\textbf{INT-4 AWQ Quantization Impact on Micro-F1 Scores.} All numbers are the percentage point difference between the micro F1 score for the FP16 model and the INT-4 AWQ quantized model on the given task.}
\label{tab:quant_results}
\end{table*}

To understand this trade-off on our tasks, we explore the performance impact of INT-4 Activation Aware Quantization (AWQ)~\cite{lin2024awqactivationawareweightquantization} on three of the Llama 3 family of models. This quantization approach reduces the memory requirement for model parameters by approximately 70\%.

As shown in \ref{tab:quant_results}, despite the large reduction in model size, we find limited impact of INT-4 AWQ quantization across Llama 3.1 8bn, Llama 3.1 70bn, and Llama 3.3 70bn. The largest performance degradation was found for Llama 3.1 8bn  with an average reduction in Micro-F1 score of 1.3\% percentage points across the 16 tasks.

\section{Discussion}

We see research developing automated evaluations of LLMs on representative public health free text, tasks, and knowledge as crucial for future successful deployment in real world use cases. This work introduces a set of these evaluations for \textit{Natural Language Understanding} tasks in order to provide an initial assessment of both LLMs' applicability in general and to compare the performance of individual LLMs. 

Overall, we find LLMs of all sizes perform strongly on the simpler public health classification tasks across all three areas (burden, risk factors, and interventions) and types of free text (academic, news, social media, and questionnaires). This is a promising sign that LLMs may already be useful tools for processing some public health text to support with real world tasks. It also demonstrates that LLMs can generate responses requiring domain-specific information about a range of public health topics, without which they could not achieve these results.

This promising performance suggests that public health professionals and LLM specialists should explore the potential benefits these models can offer in controlled deployments on simpler tasks. We see significant opportunities for LLMs to systematically structure free text, converting vital public health information embedded in text into novel structured datasets. Additionally, in situations where the sheer volume of data renders manual review impractical, LLMs could provide a scalable solution, especially during events like pandemics where data volumes grow exponentially along with case counts. Furthermore, LLMs could be employed to enhance the quality assurance of existing manual annotation processes, as a particularly low-risk way to incorporate this technology. 

However, for some public health tasks where very specific knowledge (\textit{Contact Type Classification}) is required, or where it requires extracting structured data from long pieces of free text (e.g \textit{BioDex Drugs Extraction}) LLMs perform poorly at all model sizes in the zero-shot setting. Therefore, continued model improvement and assessment may be needed before LLMs can reliably support human experts on advanced public health text processing tasks.  

These evaluations also highlight a number of specific considerations for LLMs within public health:

\textbf{Gains from advanced prompting.} For some of the hardest tasks in the evaluation, such as \textit{Contact Type Classification} and \textit{Health Causal Claims Classification}, we find few-shot prompting significantly improves performance. The literature~\cite{zhang2023potential, Chen_2024, bisercic2023interpretable} also suggests potential further gains from Chain-of-Thought (CoT) prompting techniques. Therefore, more complex LLM pipelines may improve the reliability of results even on these more challenging tasks.

\textbf{Weaker long form extraction performance.} The weakest performance across LLMs is observed on long document extraction tasks, such as \textit{BioDex Drugs Extraction}. A particular challenge of these tasks is the LLM identifying the correct span of text to extract along with applying the correct definition of the target. We find this often results in the LLM extracting too many spans or the LLM extracting the incorrect sized span for the ground-truth label.  

\textbf{Variable benchmark applicability.} We find some public LLM benchmarks, which use data that often was not explicitly designed for public health evaluation, have some limitations. A particular common issue is the lack of a well defined annotation protocol (often because the data was annotated for a different purpose). This means a prompt may lead an LLM to generate labels following a different "definition" of the labels or tasks than those used when the data was originally annotated. This may lead to LLMs that are potentially capable of performing a task generating incorrect labels. While reviewing outputs of LLMs we also noted some potentially anomalous ground truth labels in some datasets.

\textbf{Output fragility.} Anecdotally, we observe weak performance is often due in part to either the LLM failing to generate the requested output format (e.g providing an explanation instead of simply "yes" or "no") or output structure (e.g outputting an invalid JSON). For example, Flan-5-xxl performs poorly on \textit{News Headline Classification} largely due to it not being able to generate consistently well-formatted JSON outputs, rather than incorrect classifications. These issues can often be solved via more advanced prompting or post processing~\cite{agrawal2022large} and so our evaluations likely underestimate what could be achieved with bespoke pipelines.  

\textbf{Best open-weight models are increasingly comparable to private models.} We find the latest Llama-3.3-70b model performs comparably to GPT-4 and GPT-4o series models on the 11 tasks assessed. This is one indication that the latest open-weight models are becoming competitive with private models for these types of classification and extraction tasks in public health.

\textbf{Limited performance loss with INT-4 quantization.} We evaluate the performance impact of quantization on three Llama 3 family models across our tasks. We find limited impact for both the 8bn and 70bn model sizes, suggesting quantization is a promising avenue for reducing the memory footprint of open-weight LLMs for our public health tasks. 

\textbf{Task specific fine-tuning approaches.} Whilst this paper focuses on evaluating LLM in-context learning capabilities, it is important to note that literature finds that for many applications specific fine-tuned models often perform competitively~\citep{doosterlinck2023biodex, Chen_2024, guo2024evaluating}.

\textbf{Benefits of domain specific annotation protocols.} Methodologically, a crucial part of successful LLM NLU evaluations is having rigorous and consistently applied annotations, particularly in complex and subjective areas such as health. Epidemiologists have significant expertise in dealing with these issues within public health. We have found drawing on this expertise to develop detailed definitions and protocols for our evaluation datasets to be valuable. We find this approach to codifying information has been helpful both to align experts providing manual annotation, and also often incorporating these definitions into the LLM prompts leads to outputs that more closely match the prompter's intention. Descriptions of our annotation approaches can be found in \Cref{annotations} and we will share more details of the protocols developed in these projects in use-case specific public health papers. 

\section{Conclusion}

Whilst the general capability of LLMs has grown rapidly~\citep{bubeck2023sparks, openai2023gpt4, medpalm2}, assessing their performance on public health specific knowledge, tasks, and free text is an important prerequisite for successful real-world applications. In this work we take a first step towards evaluating and understanding the applicability of LLMs to public health via a broad range of automated LLM evaluations on representative tasks and anonymised or synthetic free-text. 

Our initial results suggest that LLMs may already be useful tools to support public health experts extract information from a wide variety of free text sources. This ability in turn can potentially support and scale public health surveillance, intervention, and research activities.    

We note that while some LLMs can process public health related text with a relatively high degree of accuracy, there is variable performance between models and across tasks. Appropriate validation is essential for any task, because even highly capable models can generate labels or extract data that do not match the prompt author's intentions. Some real-world applications are also likely to need assessments of context specific risks for a given LLM, task, and dataset combination, as well as approaches to testing known limitations such as output fragility or bias.

Future research is needed particularly to understand LLMs' applicability to more complex long form public health generation tasks, as well as evaluate the performance of fine-tuned domain or task specific LLMs. We also aim to continue extending public health automated evaluations for novel types of tasks and data, as well as new LLMs.
\section*{Acknowledgements}

This work was enabled by UKHSA HPC Cloud \& DevOps Technology colleagues developing and maintaining internal HPC resources. We would also like to thank our UKHSA colleagues in Clinical and Public Health for their support and expertise in developing this work. 

\section*{Ethical Considerations}

To avoid potential detrimental impacts of text processing on individuals, we employ only public, synthetic, or anonymous data, with some datasets spanning multiple categories. We also prioritise research on less sensitive data sources and lower risk tasks, especially in comparison to others in healthcare settings.

It is important to mitigate potential risks within public health text processing in order to avoid errors causing harm to individuals or communities. In this paper we have provided initial results to assist researchers in understanding potential LLM performance on public health tasks. This assessment is important, because it helps set out the potential opportunities to use LLMs to generate public health insight and support public health action; however, future application should be done with sufficient mitigation measures in place. 

In terms of mitigation, for real-world deployments it is imperative that any LLM-based software be evaluated within its specific context. The results reported herein are not sufficient to endorse the deployment of software for these public health tasks, without in-context assessments. For further discussion of these issues and other wider evaluation that is important to consider for real-world deployments see \Cref{lit_outcome}.

\clearpage
\bibliographystyle{unsrtnat}
{\small{\bibliography{refs}}}

\clearpage

\section{Appendix: Literature on Domain Specific Evaluations}\label{lit_review_full}

\begin{figure}[h]
    \centering
    \includegraphics[scale=0.6]{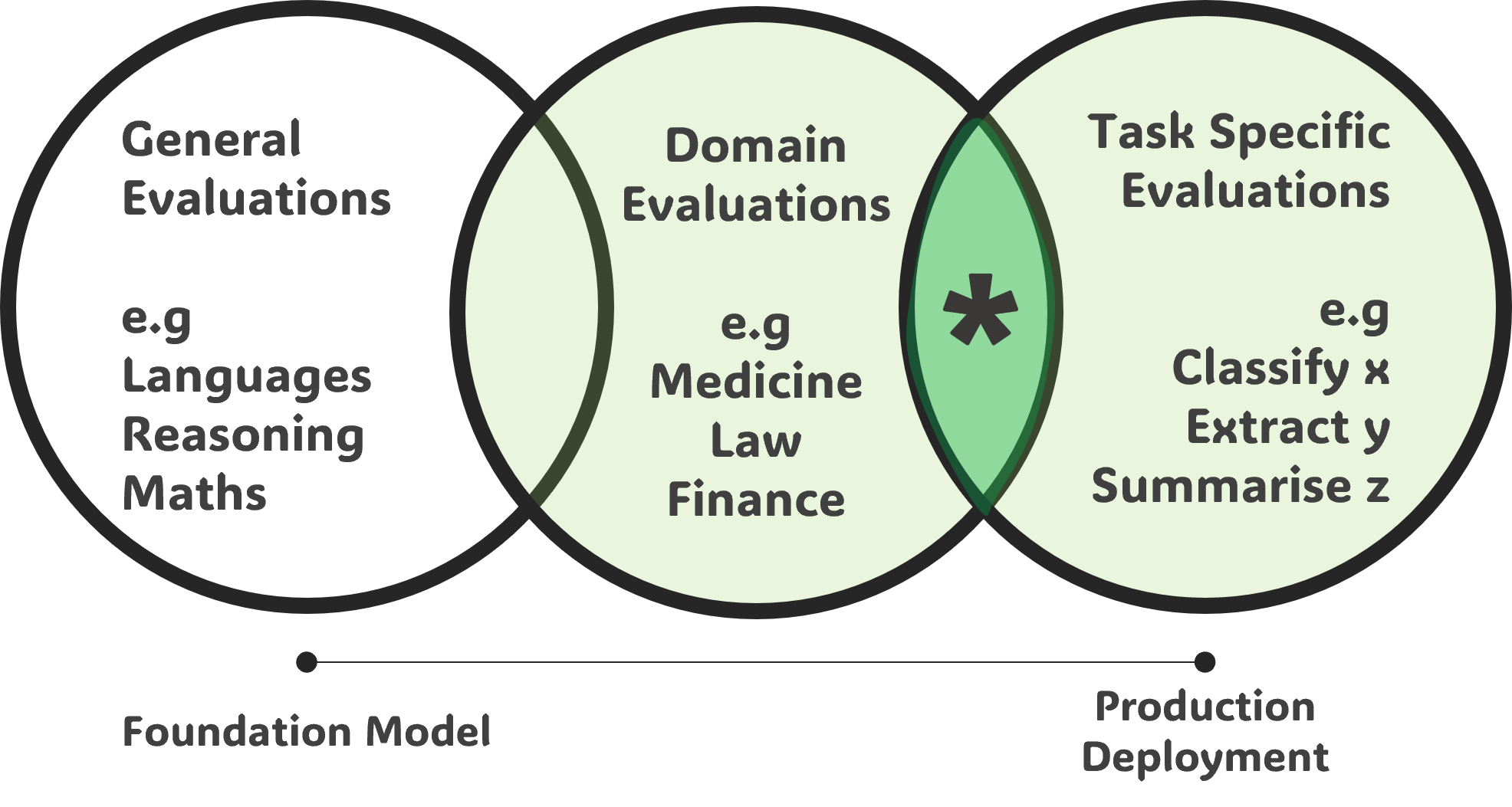} 
    \caption{\textbf{LLM Evaluation Spectrum.} In this paper we focus on a combination of domain and task specific LLM evaluations within public health in order to inform our understanding of where LLMs may be successfully deployed within the field. The area marked by the \textbf{(*)} - denotes how our evaluations compare to others.} \label{fig:venn_eval_fig}
\end{figure}

Evaluation of LLMs is a broad and rapidly growing field of research \cite{chang2023survey} ranging from very general assessments of capabilities or intelligence to very task specific performance results (\Cref{fig:venn_eval_fig}). 

However, while researchers have developed evaluations for related fields, there is limited literature assessing open-weight LLMs in a standardised way across a variety of public health specific tasks and data.

\subsection{Approaches to Domain Specific LLM Evaluations}

Domain specific evaluations of LLMs can take a number of forms depending on several factors: (1) level of generality, (2) types of task, and (3) outcome of interest.

\subsubsection{Level of Generality}\label{lit_generality}

Domain specific evaluations often lie on a spectrum, from assessing the potential applicability of an LLM for a domain in general (e.g Medicine \cite{medpalm2}) to assessing an LLM on a single specific task within a domain (e.g diagnosing neuro-ophthalmic diseases \cite{madadi2023chatgpt}). 

General domain evaluations often draw on existing human assessments and exams. For example, in Medicine using questions from (or in the style of) the US Medical Licensing Examination (USMLE)~\citep{chatgptUSMLE, lievin2022can, medpalm, chen2023meditron70b}, in Law the US Bar Exam~\citep{bommarito2022gpt, katz2023gpt}, or using human exams for given subjects~\citep{openai2023gpt4, chatgptAKT, bommarito2023gpt, choi2023chatgpt}. However, bespoke domain evaluations for LLMs are increasingly being developed (often including aspects of the human assessments), such as LawBench (Legal)~\cite{fei2023lawbench}, MultiMedQA (Medical)~\cite{medpalm}, FinBen (Financial)~\cite{xie2024finben}, the Financial Language Understanding
Evaluation (FLUE) benchmarks (Financial)~\cite{shah2022flue}, and ChemLLMBench (Chemistry)~\cite{guo2023large}.

In contrast, the more specific domain evaluations in the literature, which target either sub-fields or specific tasks, have often involved collecting and manually annotating new evaluation datasets (or modified / filtered existing domain specific data). In Medicine this has commonly involved collecting data and evaluating LLMs for specific sub-fields, including, Neuro-ophthalmology (diagnosis)~\cite{madadi2023chatgpt}, Bariatric surgery (QA)~\cite{samaan2023assessing}, Osteoarthritis (case management)~\cite{chen2024evaluating}, Dementia (diagnosis)~\cite{wang2023llms}, and Genetics (QA)~\cite{duong2023analysis}. In Law, specific evaluations have generally been carried out for given tasks or abilities, including, generating explanations for legal terms~\cite{savelka2023explaining}, statutory reasoning~\cite{blair2023can}, and legal entailment~\cite{yu2022legal}.

The chosen level of generality for an evaluation is primarily determined by the overall aim. More general domain evaluations are most useful for understanding broad capabilities, developing or fine-tuning new LLMs, and prioritising existing LLMs for further assessment. More task specific domain evaluations are generally crucial when considering deploying LLMs for real world use cases or applying LLMs to unseen datasets and tasks (such as private organisational data).

\subsubsection{Types of Task}\label{lit_task_type}
In the literature, the approach adopted for a domain specific LLM evaluation is also significantly influenced by the type of task (or tasks) involved. As discussed by \citet{chang2023survey}, a key factor is whether the task primarily involves classification or inference (Natural Language Understanding - NLU) or whether it involves generating free text (Natural Language Generation - NLG).

Evaluation of LLMs for tasks focusing on classification or the extraction of structured data (NLU) utilises similar metrics and approaches to traditional data science. This involves collecting a representative dataset of free text, annotating with ground truth labels, and then evaluating the performance of the LLM using metrics such as accuracy, recall, precision, and F1 scores~\citep{doosterlinck2023biodex, agrawal2022large, shyr2023identifying}.

In contrast, LLM tasks that generate unstructured free text (NLG), such as summarisation, have been shown to be hard to evaluate with traditional automated methods~\citep{tang2023evaluating, goyal2022news, chang2023survey, chen2024large}, such as ROUGE~\cite{lin2004rouge}. This has led many NLG task evaluations to adopt human evaluation approaches, particularly in domain specific or risk-averse fields, such as Medicine~\citep{medpalm, medpalm2, shaib2023summarizing, chen2024large}. This usually involves human experts reading the LLM outputs and scoring them on absolute (rate out of 10) or relative (which response is better) metrics for a given criteria.

However, human evaluation of NLG tasks brings a number of challenges, including: (1) cost, as expert annotation is time-consuming and experts' time is valuable, (2) subjectivity, as there may be inconsistency in annotations between experts, and (3) scalability, as human evaluations must be repeated manually for every model (whereas annotated data can be used to evaluate all models). These issues, combined with the increasing capabilities of the state of the art LLMs, have led researchers to investigate replacing manual human review of generation tasks, with automated review by different, usually more capable LLMs~\citep{zheng2023judging, bai2023benchmarking, chiang2023large}. Whilst these approaches have seen some success~\citep{zheng2023judging, chiang2023large} it remains unclear which domains or tasks LLMs are \textit{"qualified"} to evaluate. This is a particular issue for risk-averse sectors that involve specialised knowledge, such as public health.

\subsubsection{Outcome of Interest} \label{lit_outcome}
The appropriate methodology for domain specific evaluations is also strongly influenced by the exact outcome researchers are interested in assessing. For NLU tasks, the potential outcomes of interest can range from those focused on test set performance (e.g accuracy), which we look at in this paper, to broader considerations such as bias and robustness. 

For automated evaluations the literature largely uses performance metrics, with a particular focus on F1 scores~\citep{zhang2023potential, guo2024evaluating, shyr2023identifying, Chen_2024, agrawal2022large}. For specific tasks these high level metrics are often supplemented with further class specific analysis~\cite{shyr2023identifying}. 

Some of the other key outcomes of interest for fields such as public health are assessments of software for bias and fairness, where this bias could be caused by bias in the LLM or other places in the input data or software. General assessments of LLM bias are often carried out during LLM training and benchmarking~\citep{touvron2023llama, parrish-etal-2022-bbq, claude3, taylor2022galactica, geminiteam2024gemini}. For specific NLU tasks, further bias evaluations often involve analysing label specific results to identify whether the predictions or errors deviate significantly based on individual characteristics or group characteristics~\citep{unintended_bias, chang2023survey, gallegos2024bias}. 

Evaluating bias and fairness is crucial for understanding the potential risks when deploying LLMs for real-world use cases, particularly those that involve text about certain communities. In a UK context, protected characteristics are enshrined in law as characteristics on which people cannot be discriminated against~\cite{equality_act_2010}. Specifically for public health, evaluations may also need to consider wider health equity frameworks, such as CORE20PLUS~\cite{ukhsa_he_board}. Evaluations of bias and fairness can generally be most effectively carried out using the contents of the free-text (focusing on the specific protected characteristics that are relevant), task (assessing the risk of bias in the output), and deployment process (evaluating the software as a whole and how a human expert in the loop could mitigate or perpetuate risks) that would be used in production.

\subsubsection{Summary}

\begin{figure}[h]
    \centering
    \includegraphics[scale=0.6]{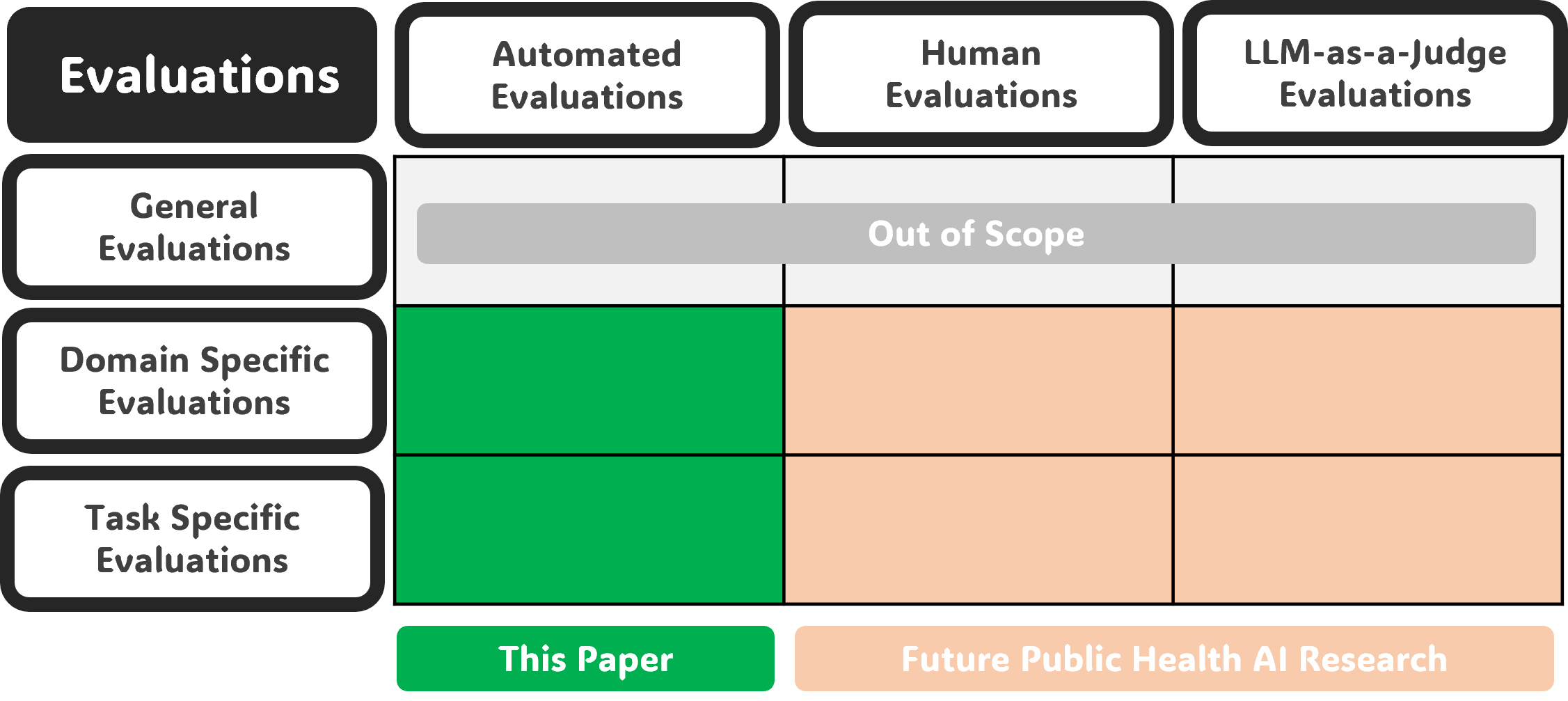}
    \caption{\textbf{Overview of Evaluations}. We focus on automated evaluations of classification and extraction tasks. }\label{fig:eval_table_fig}
\end{figure}

Overall, our aim with this work is to provide initial assessments of LLMs within public health. To cover the broadest range of models and areas, we focus on automated domain specific evaluations across a range of relevant NLU tasks, see \Cref{fig:eval_table_fig}.

\subsection{Existing Evaluations of LLMs in Public Health and Related Fields}\label{lit_evals}

The literature contains a number of existing domain specific NLU LLM evaluations that are relevant for public health, some of which we adopt relevant sub-sets of within our evaluations.

\subsubsection{Disease and Health Issue Classification and Extraction}\label{disease_lit}

Classifying and extracting potential diseases and health issues from free text is an important task within public health analysis and has significant overlap with the Medical domain. 

\citet{zhang2023potential} evaluate ChatGPT (GPT-3.5) and GPT-4 at the single label task of disease classification from electronic health records (EHRs). GPT-4 with zero-shot prompting was found to achieve between 0.75 and 0.96 F1 score across the 5 diseases and outperformed GPT-3.5 on 4 of the 5 diseases. Sensitivity analysis suggested few-shot with Chain of Thought (CoT) prompting improved GPT-4's performance. 

Similarly, \citet{guo2024evaluating} evaluate GPT-4 and GPT-3.5 (along with fine-tuned classifiers) across six single label health information classification tasks using manually annotated Twitter (now called X) data and data from the Social Media Mining for Health Applications (SMM4H) datasets. As found in other work, GPT-4 outperforms GPT-3.5 using zero-shot prompting but with a high variance across tasks, with F1 scores ranging from 0.35 and 0.8. 

Finally, due to the importance of novel pathogens within public health, the literature dealing with rare diseases (where information in the LLM pre-training data is likely to be more limited) is particularly interesting for LLM evaluation. 

\citet{chen2024rarebench} introduce RareBench to evaluate LLMs' (including GPT-4, GPT-3.5, Gemini, Llama-2-7b and Mistral-1-7b) ability to perform rare disease phenotype extraction, screening, and diagnosis. GPT-4 is found to have the highest performance across all tasks using zero-shot prompting while the smaller 7b parameter open-weight Llama-2 and Mistral-1 models have the lowest scores.

\citet{shyr2023identifying} evaluate ChatGPT for extracting rare diseases, signs and symptoms using the the RareDis corpus~\cite{martínezdemiguel2021raredis}. ChatGPT with zero-shot prompting and relaxed matching achieves overall F1 scores of 0.41 to 0.47, but with significant variation across types from 0.15 (symptoms) to 0.76 (rare diseases). Few-shot prompting was generally found to lead to better performance. The complexity of the labels meant that evaluation approaches requiring exact matching of outputs and labels had significantly worse results.

\subsubsection{Drug and Intervention Extraction}\label{drug_int_lit}

Extracting relevant Pharmaceutical interventions from free text is another common task in public health free text processing.

\citet{doosterlinck2023biodex} propose BioDEX, a dataset of PubMed articles about Adverse Drug Events (ADE) and their corresponding ground truth ADE extractions. BioDEX is used to evaluate GPT-3.5 and GPT-4 LLMs with few-shot prompting, this achieves an  overall F1 score of approximately 0.5 at extracting the 4 core ADE report attributes (patient sex, serious event, reactions and drugs), currently well below expert level. Due to the length of the papers the few-shot prompts are constructed using abstracts, and the input paper is truncated to fit the context window, potentially limiting the information available to the model.

\citet{agrawal2022large} add new annotations to the Clinical Acronym Sense Inventory (CASI) dataset~\cite{moon2014sense} to evaluate GPT-3 for the task of extracting medical interventions from clinical notes and medical acronym disambiguation. Combining GPT-3 with a \textit{resolver} to form structured outputs along with zero-shot and one-shot prompting is shown to have fairly strong results across tasks, generally achieving F1 scores >0.6. However, as in other studies, the authors note the difficulty of generating exact matches for complex ground truth labels.  

Similarly, \citet{bisercic2023interpretable} evaluate InstructGPT for the task of generating structured JSON data from medical reports within their TEMED-LLM approach. Using medical reports on patient treatments, psychologist notes, strokes, hepatitis, and heart disease, InstructGPT with one-shot and CoT prompting combined with additional output validation (e.g correcting JSON formatting errors) achieves over 90\% accuracy for 4 / 5 tasks and outperforms one-shot prompting.

Finally, identifying and understanding relevant medical and health recommendations is a key capability for interpreting public health guidance. Related to this, \citet{Chen_2024} evaluate GPT-4 and GPT-3.5 on single label Biomedical classification tasks using the HealthAdvice~\cite{li2021EMNLPHealthAdvice} and CausalRelation~\cite{yu2019EMNLPCausalLanguage} datasets. The HealthAdvice dataset is used to evaluate the LLMs' ability to identify whether a given sentence contains no, weak, or strong health advice. The CausalRelation reasoning evaluation tests the LLMs' ability to identify casual and correlation claims in PubMed conclusion sentences. Testing a range of prompts from zero-shot to few-shot with CoT, GPT-4 is shown to slightly outperform GPT-3.5 with 0.65-0.77 macro-F1 scores across the tasks using the highest performing prompt (few-shot with CoT).

\subsubsection{Relevant Sub-sets of General Evaluations}\label{sub_set_lit}

In addition to these domain specific evaluations in the literature, it is also important to note that there are subsets of general LLM evaluations that are relevant for public health (although detailed breakdowns of subset results often are not reported). 

The Massive Multitask Language Understanding (MMLU) benchmark~\cite{mmlu} has subsets covering general medical and scientific knowledge. More importantly, it also contains subsets relating to virology, genetics, and nutrition, that are particularly relevant for some public health sub-fields. 

Also relevant for public health is the PubMedQA~\cite{jin2019pubmedqa} Biomedical question answering benchmark now included in the broader MultiMedQA~\cite{medpalm} medical benchmark. The labelled subset of PubMedQA consists of 1000 PubMed articles that have questions in the title, combined with the manually annotated answer (yes, no, maybe) to the question based on a review of the abstract (with and without concluding sections). This is particularly relevant for assessing the ability of LLMs to infer conclusions to scientific questions based on a summary of the evidence.   

More recently, the Massive Multi-discipline Multimodal Understanding and Reasoning (MMMU)~\cite{yue2023mmmu} benchmark has a specific subset for public health. However, due to the multi-modal nature of the benchmark it currently sits outside the scope of this paper. 

\normalsize

\clearpage

\section{Appendix: Annotations}\label{annotations}

 Seven of the datasets we use to evaluate LLM performance are datasets annotated in collaboration with public health specialists in the relevant sub-fields of public health. Given the comprehensive nature of this evaluation, the detail of the exact approach to manual annotation is left for future public health specific papers. However, this appendix provides a summary of the level of rigour used to ensure high quality annotations that would be reproducible by other human experts (in order to give LLMs a chance at producing similar results).

\begin{table*}[htbp]
\footnotesize
\centering
\begin{tabularx}{\textwidth}{p{2cm} p{1.5cm} p{5cm} p{4cm}} 
\toprule
\textbf{Dataset} & \textbf{Origin} & \textbf{Annotation approach} & \textbf{Annotation confidence} \\
\midrule
Gastrointestinal Illness Classification & Bespoke for LLM evaluation & A forthcoming publication will set out the methodology in more detail. A detailed protocol was honed by several public health specialists. The protocol was independently applied by at least two reviewers on a sample of 200. This confirmed greater than 90\% agreement between reviewers, and each discrepancy was re-reviewed and resolved by agreement between all reviewers. The rest of the 3000 sample was assigned to a single reviewer who had the opportunity to check uncertain examples with other reviewers. & User-generated online restaurant reviews contain highly variable information, making it hard to apply a protocol. However, the considerable degree of consistency between human reviewers gives more confidence. \\

\midrule Gastrointestinal Illness Symptom Extraction & Bespoke for LLM evaluation & A forthcoming publication will set out the methodology in more detail. A detailed protocol was honed by several public health specialists. The protocol was then applied by a reviewer involved in the GI Classification task, uncertain classifications were shared with a second reviewer to check. & User-generated online restaurant reviews contain highly variable information, making it hard to apply a protocol. However, the considerable degree of consistency between human reviewers gives more confidence. \\

\midrule ICD-10 Description Classification & Pre-existing organisational & A forthcoming publication will set out the methodology in more detail. The assignment of annotations is based on: public health experts in the a given disease area, review of the literature, and other pre-existing labels by public health experts. & Some of the labels in this dataset are assigned trivially, where a description clearly relates to an infection. Other labels, where a disease is labelled infection related, because the disease has an infectious aetiology can be considerably harder. Considerable validation with subject matter experts and triangulation between approaches leads to considerably higher confidence. \\

\midrule News Headline Classification & Bespoke for LLM evaluation & A detailed protocol was honed by several public health specialists. A single reviewer then assigned labels based on the protocol, spot checking was done by another reviewer who confirmed a high degree of agreement. & The headlines in this work are relatively easy to annotate. However, using a single reviewer with spot checking is a less rigorous process than the double blind review used for some of the other datasets. \\ 

\bottomrule
\end{tabularx}
\caption{\textbf{Manual Annotation.} Summary of the approach to manual annotation for each task.}
\end{table*}
\clearpage
\begin{table*}[ht!]
\footnotesize
\centering
\begin{tabularx}{\textwidth}{p{2cm} p{1.5cm} p{5cm} p{4cm}} 
\toprule
\textbf{Dataset} & \textbf{Origin} & \textbf{Annotation approach} & \textbf{Annotation confidence} \\

\midrule Contact Type Classification & Bespoke for LLM evaluation & A detailed protocol was honed by several public health specialists. The data was generated by GPT-4, which was prompted to generate text and assign a label on generation. Two independent reviewers then also manually annotated their label. Over 90\% agreement was confirmed and any discrepancies reconciled.  & The independent annotation by two human experts and GPT-4, where discrepancies were resolved, leads to high confidence that the protocol (also provided to the LLM in the prompt) was followed. \\ 

\midrule Country Disambiguation & Pre-existing organisational & Country labels were manually assigned to ambiguous free text (that could not be matched to a country through key word searches) by single reviewers, with spot checking by a second reviewer. The approach to assigning labels also involved reviewers using search engines to try to ascertain the country of sub-country geographical units mentioned in the free text. & Some of the labels are inherently uncertain (for instance, the text only refers to a location that exists in several countries, or a spelling mistake like Nigeri means Nigeria and Niger are both possible countries). However, spot checking reveals a low error rate, with disagreements in labels largely relating to inherent uncertainty. \\
\midrule
Food Extraction & Bespoke for LLM evaluation & A forthcoming publication will set out the methodology in more detail. A detailed protocol was honed by several public health specialists. The protocol was then applied by a reviewer involved in the GI Classification task, uncertain classifications were shared with a second reviewer to check. & User-generated online restaurant reviews contain highly variable information, making it hard to apply a protocol. However, the considerable degree of consistency between human reviewers gives more confidence. \\ 

\midrule Guidance Topic Classification & Pre-existing organisational & The Topic labels were assigned based on which team in UKHSA wrote the piece of guidance. Pieces of guidance were included only if teams have a clear remit in a specific sub-field of public health (e.g. Sexual Transmitted Infections) rather than an organisational remit that would be less relevant to LLMs (e.g. Health Protection Operations). & Teams in UKHSA only publish guidance within their team remit, so there is a very high degree of confidence in the labels assigned. Some pieces of guidance are more uncertain where they related to topics that fall at the intersection of two teams' remits.\\

\midrule Guidance Recommendation Classification & Bespoke for LLM evaluation & A detailed protocol was honed by several public health specialists. The protocol was applied by at least two reviewers for a sample of 100 examples. >90\% agreement was confirmed before a single reviewer was assigned the remaining labels with the option to send uncertain labels to a second reviewer. & Recommendations are an abstract concept,  which makes them more challenging to define. However, the detailed protocol and considerable inter-reviewer agreement leads to more confidence that the annotation is correct. \\
\bottomrule
\end{tabularx}
\caption{\textbf{Manual Annotation continued.} Summary of the approach to manual annotation for each task}
\end{table*}

\clearpage

\section{Appendix: Example Prompts}

\begin{figure}[H]
    \centering
    \includegraphics[scale=0.6]{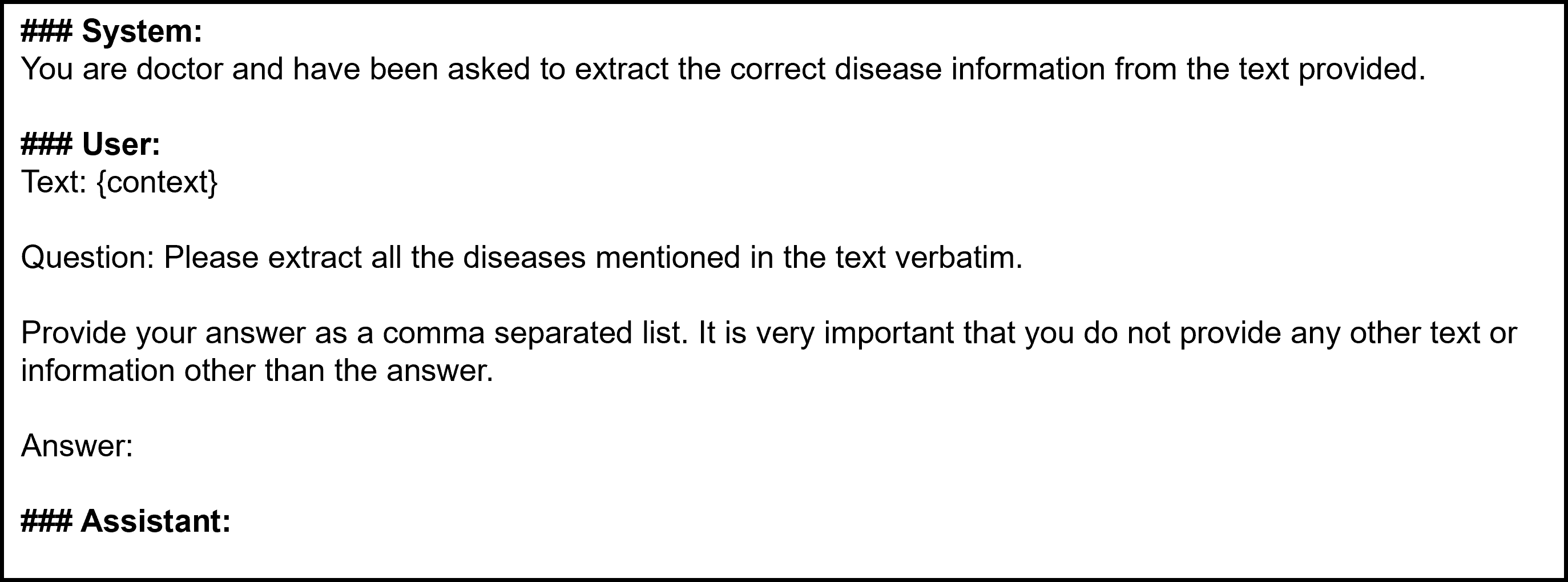}
    \caption{\textbf{Example Zero Shot Prompt for NCBI Disease Extraction}. This is an example of the zero shot prompt used in the \textit{NCBI Disease Extraction} task structured with the Stable-Beluga-2 prompt template.}\label{fig:prompt_zero_ncbi}
\end{figure}

\end{document}